\documentclass[11pt]{article}
\usepackage[utf8]{inputenc}

\usepackage{color}
\usepackage{xcolor} 
\usepackage{listings} 
\usepackage{acl}

\usepackage{CJKutf8}

\usepackage{times}
\usepackage{latexsym}

\usepackage[T1]{fontenc}

\usepackage[utf8]{inputenc}

\usepackage{amsmath}
\usepackage{graphicx}
\usepackage{booktabs}
\usepackage{multirow}
\usepackage{pifont}

\usepackage{microtype}

\usepackage{inconsolata}

\newsavebox\CBox
\def\textBF#1{\sbox\CBox{#1}\resizebox{\wd\CBox}{\ht\CBox}{\textbf{#1}}}

\title{YAYI 2: Multilingual Open-Source Large Language Models}

\author{
Yin Luo$^{*}$, Qingchao Kong$^{*}$, Nan Xu$^{*}$, Jia Cao, Bao Hao, Baoyu Qu, Bo Chen, Chao Zhu\\
Chenyang Zhao, Donglei Zhang, Fan Feng, Feifei Zhao, Hailong Sun, Hanxuan Yang, Haojun Pan\\
Hongyu Liu, Jianbin Guo, Jiangtao Du, Jingyi Wang, Junfeng Li, Lei Sun, Liduo Liu, Lifeng Dong\\
Lili Liu, Lin Wang, Liwen Zhang, Minzheng Wang, Pin Wang, Ping Yu, Qingxiao Li, Rui Yan, Rui Zou\\
Ruiqun Li, Taiwen Huang, Xiaodong Wang, Xiaofei Wu, Xin Peng, Xina Zhang, Xing Fang, Xinglin Xiao\\
Yanni Hao, Yao Dong, Yigang Wang, Ying Liu, Yongyu Jiang, Yungan Wang, Yuqi Wang\\
Zhangsheng Wang, Zhaoxin Yu, Zhen Luo, Wenji Mao, Lei Wang$^{*}$, Dajun Zeng$^{*}$ \\
  \textbf{Beijing Wenge Technology Co., Ltd.} \\
  \textbf{Institute of Automation, Chinese Academy of Sciences}
  }

\begin{document}

\maketitle
\renewcommand{\thefootnote}{\fnsymbol{footnote}}
    \footnotetext[1]{Corresponding authors: \{$\mathrm{yin.luo, nan.xu, lei.wang\}}$\\$\mathrm{@wenge.com}$, $\mathrm{\{qingchao.kong, dajun.zeng\}@ia.ac.cn}$.}

\renewcommand{\thefootnote}{\arabic{footnote}}
\thispagestyle{plain} 
\begin{abstract}
As the latest advancements in natural language processing, large language models (LLMs) have achieved human-level language understanding and generation abilities in many real-world tasks, and even have been regarded as a potential path to the artificial general intelligence. To better facilitate research on LLMs, many open-source LLMs, such as Llama 2 and Falcon, have recently been proposed and gained comparable performances to proprietary models. However, these models are primarily designed for English scenarios and exhibit poor performances in Chinese contexts. In this technical report, we propose YAYI 2, including both base and chat models, with 30 billion parameters. YAYI 2 is pre-trained from scratch on a multilingual corpus which contains 2.65 trillion tokens filtered by our pre-training data processing pipeline. The base model is aligned with human values through supervised fine-tuning with millions of instructions and reinforcement learning from human feedback.
Extensive experiments on multiple benchmarks, such as MMLU and CMMLU, consistently demonstrate that the proposed YAYI 2 outperforms other similar sized open-source models.
 
\end{abstract}

\section{Introduction}

Large language models (LLMs) \cite{vaswani2017attention, kaddour2023challenges} have shown strong capabilities in language understanding and comprehension \cite{brown2020language}, as well as in common-sense Q\&A, programming and logical reasoning  \cite{lightman2023let}. Since the launch of ChatGPT, a large number of LLMs have been proposed by different institutions and companies around the world, which mostly serve as intelligent personal assistants through a chat interface, and excel at creative writing, summarizing texts, planning activities, etc. Due to the comprehensive capabilities, LLMs are even regarded as a potential path towards the artificial general intelligence (AGI).

Terabytes of training data and expensive computing resources have become the main bottlenecks restricting the development of LLMs. Several representative LLMs-based products such as ChatGPT and Claude~\cite{bai2022constitutional} are closed-source models. To make it more accessible for researchers, many open-source LLMs have been proposed. For example, BLOOM~\cite{workshop2022bloom} is the first multilingual LLM with 175 billion parameters trained on the ROOTS corpus.
Llama~\cite{touvron2023llama, touvron2023llama2} series models have achieved comparable performances with GPT-3.5 and Palm 2~\cite{anil2023palm} by training on more text tokens with better quality. Besides the ROOTs corpus, more datasets such as RedPajama~\cite{redpajamav2} and RefinedWeb~\cite{penedo2023refinedweb} are open-sourced to further facilitate LLMs training. However, these open-source datasets contain only a small portion of Chinese text and lack the common knowledge and culture about China, which severely limits the applications of open-source LLMs in Chinese-related scenarios. To fill this gap, several Chinese-based LLMs are proposed, including ChatGLM~\cite{zeng2022chatglm}, Baichuan 2~\cite{baichuan2} and Qwen~\cite{bai2023qwen}. 

\begin{figure*}
  \centering
      \includegraphics[width=0.88\textwidth,trim= 0 0 0 0]{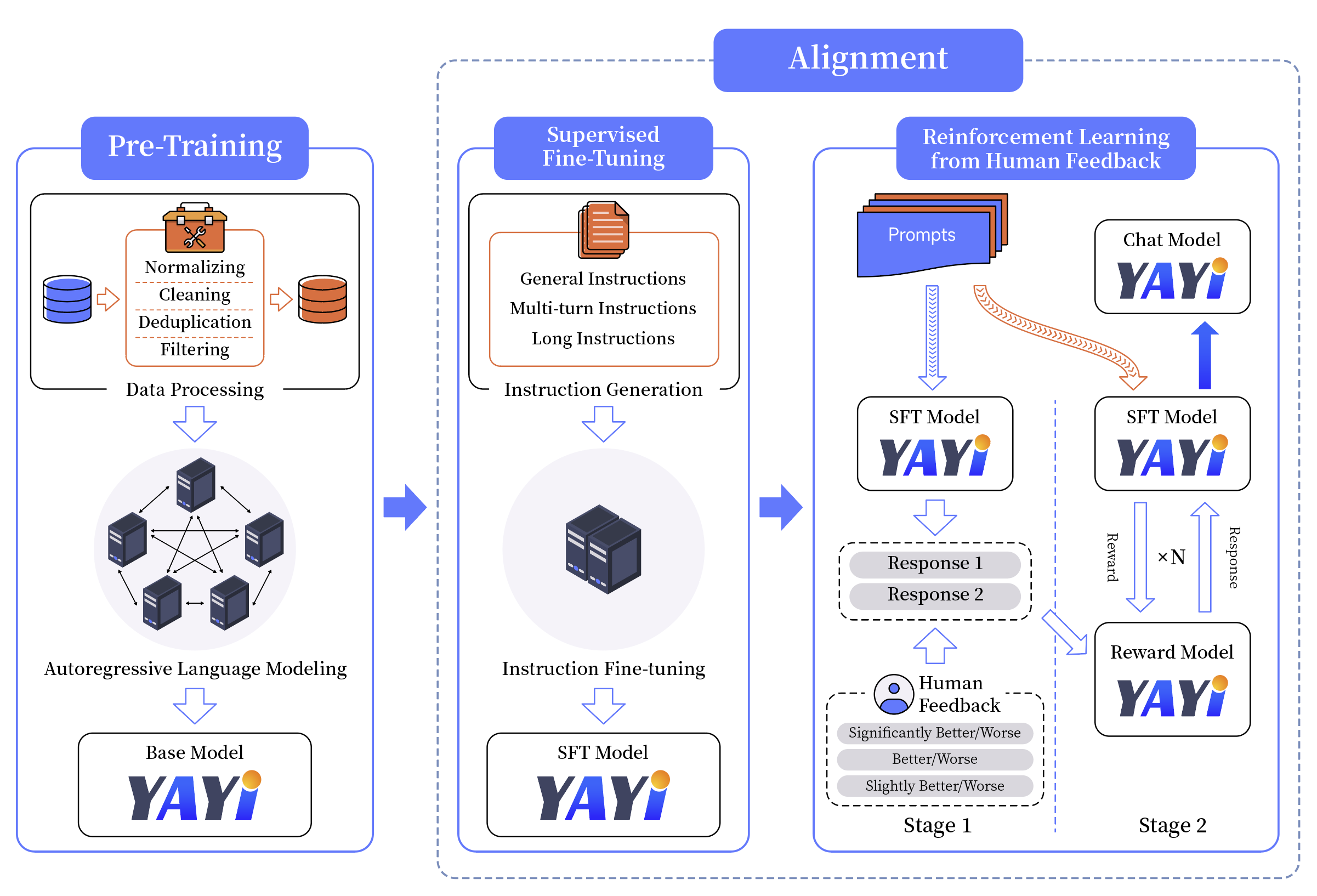}
  \caption{Training procedure of YAYI 2 base and chat models.}
  \label{fig:whole_architecture}
\end{figure*}

In this technical report, we propose a series of multilingual LLMs, denoted as YAYI (
\begin{CJK*}{UTF8}{gbsn}{雅意}
\end{CJK*}) 2, including base and chat models, both with 30 billion parameters. YAYI 2 models are trained on 2.65 trillions tokens on a computing cluster of over 1000 A800 GPUs. For pre-training dataset, we collect over 240 terabytes of texts, including news, books, Wikipedia, code, etc., of which 41.5\% are Chinese. In addition, we design a rigorous pre-training data processing pipeline, consisting of normalizing, heuristic cleaning, multi-level deduplication, and toxicity filtering. To speed up the training and inference speed, the FlashAttention 2~\cite{dao2023flashattention2} and multi-query attention (MQA)~\cite{Shazeer2019FastTD} are adopted. We elaborate the training details and optimizing techniques to  improve the training efficiency. We align YAYI 2 base model through supervised fine-tuning (SFT) with millions of instruction-output pairs and reinforcement learning from human feedback (RLHF), with better support for long instructions and multi-turn conversations. The training procedure of YAYI 2 base and chat models are shown in Figure~\ref{fig:whole_architecture}. We conduct comprehensive experiments to evaluate the effectiveness of the proposed base model. The experimental results show that the proposed model outperforms other similar-sized open-source LLMs on benchmarks covering knowledge understanding, math reasoning, and programming, and even demonstrates superiority on some benchmarks over the LLM with much larger parameters.



\section{Pre-Training}
This section provides details on the pre-training process from four aspects: pre-training data, tokenization, model architecture, and training details. We first summarize the sources of pre-training data and propose a self-developed data processing pipeline. Leveraging high-quality cleaned data, we construct YAYI 2 multilingual tokenizer. Next, we elaborate the model architecture and parameter settings. Finally, we introduce the computing cluster configuration and training strategies along with some model training tricks.

\subsection{Pre-Training Data}

\begin{figure*}
  \centering
      \includegraphics[width=0.95\textwidth]{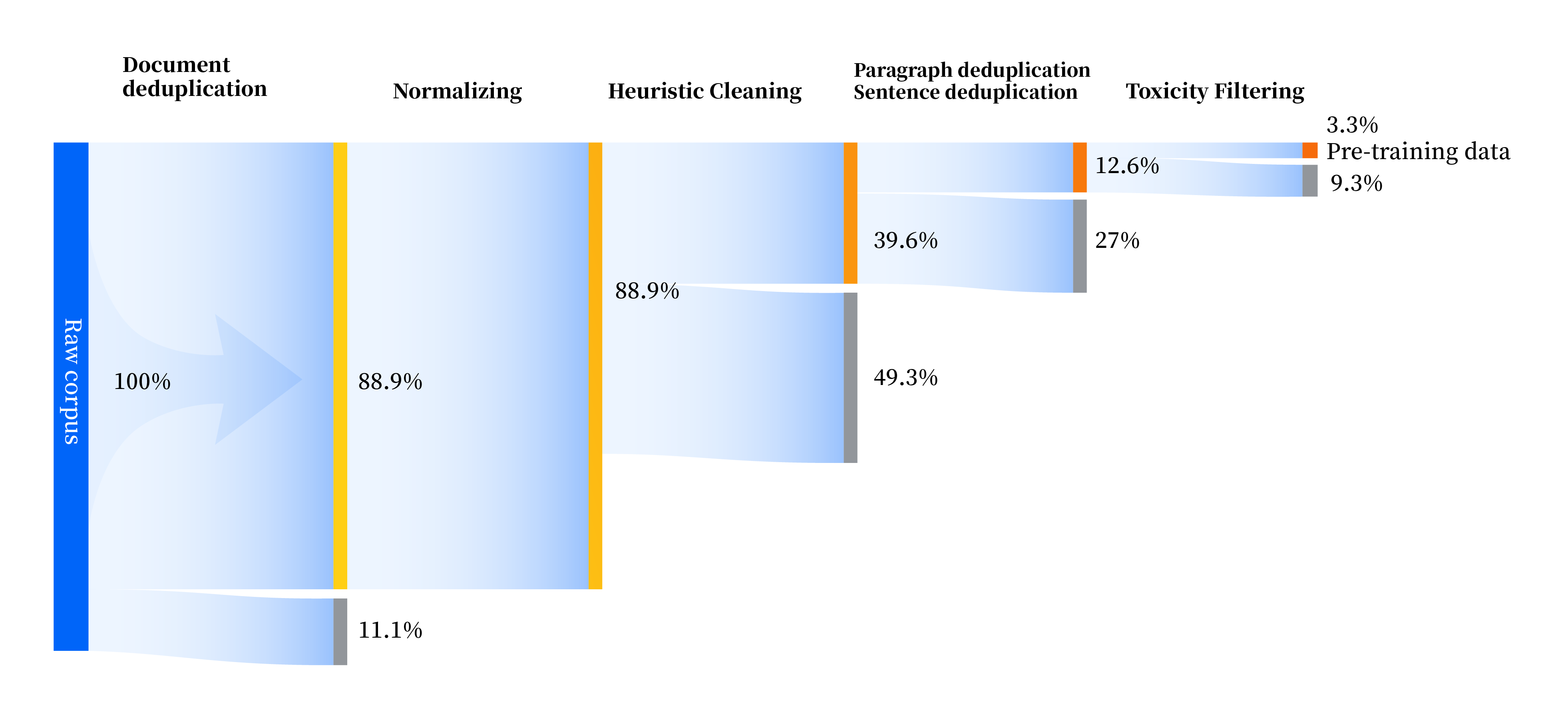}
  \caption{Pre-training data processing pipeline.}
  \label{fig:data_process}
\end{figure*}

\subsubsection{Data Distribution}

\begin{figure}
  \centering
      \includegraphics[width=0.58\textwidth, trim = 280 0 10 0,clip]{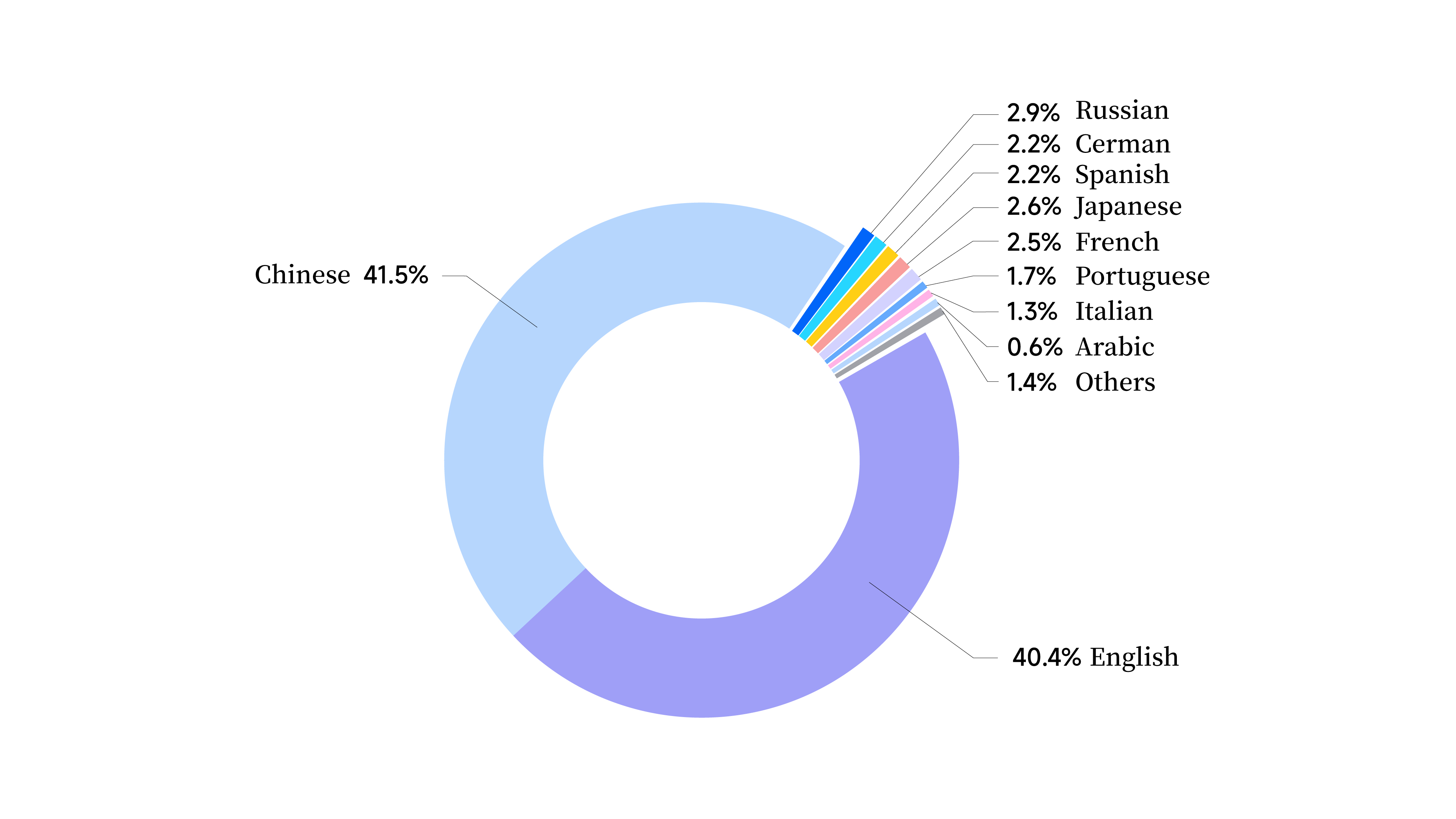}
  \caption{Language distribution.}
  \label{fig:language_distribution}
\end{figure}

The objective of pre-training is to accumulate a wide range of knowledge from all over the world and acquire various professional proficiency such as math, coding, and logical reasoning, which should give the model's responsiveness to multilingual scenarios and diverse data formats. In pursuit of the above goals, a large amount of internet data is used to train the language understanding and expression capabilities, supplemented by curated general data and domain-specific data to further enhance professional skills of the model. Figure~\ref{fig:data_distribution}\&\ref{fig:language_distribution} show the distributions of data categories and languages, respectively. Details of the data distribution are as follows:

\begin{figure}
  \centering
      \includegraphics[width=0.6\textwidth,trim = 280 0 10 0, clip]{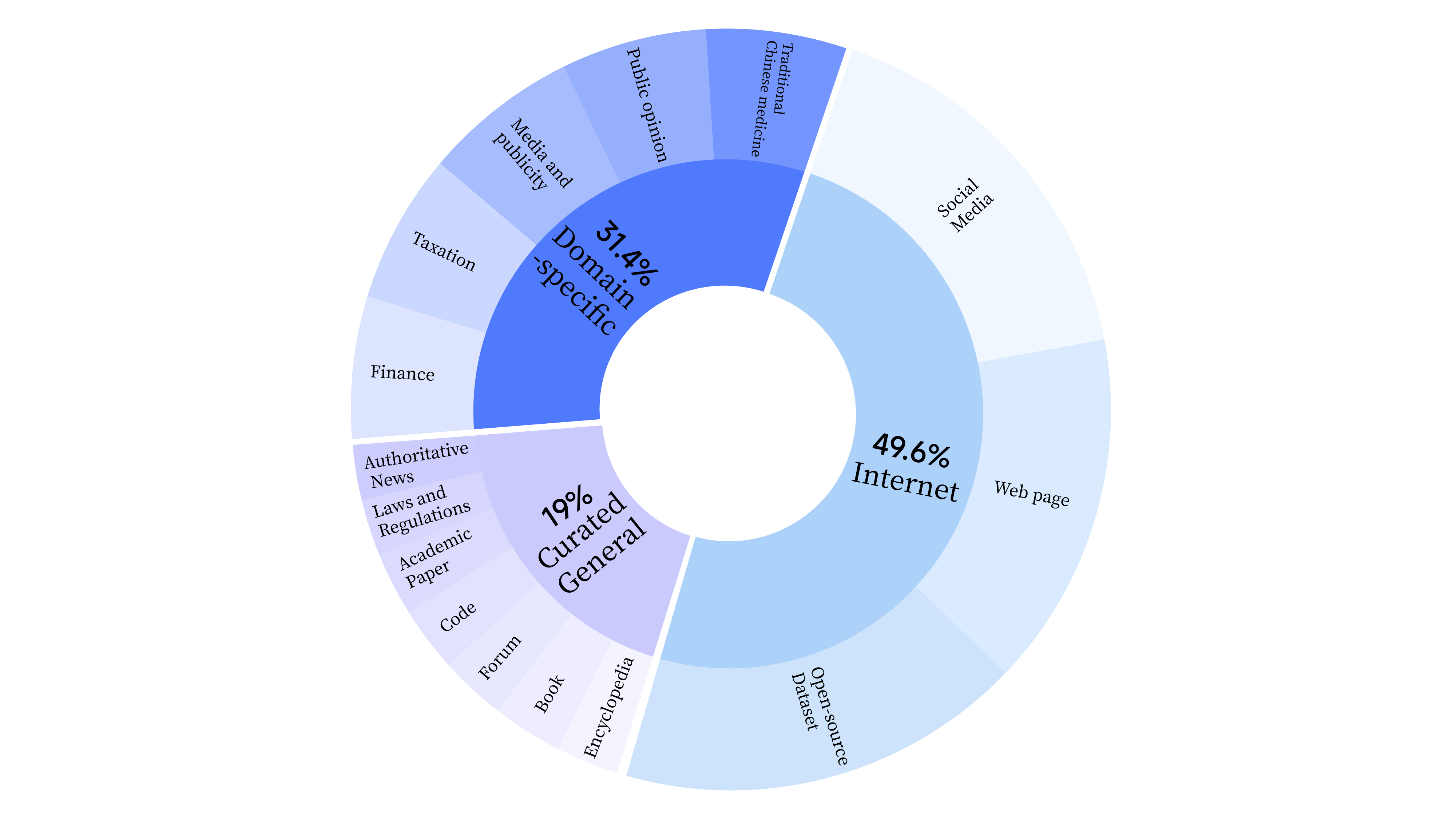}
  \caption{Data distribution.}
  \label{fig:data_distribution}
\end{figure}

\begin{itemize}
    \item Internet data primarily comprises private data consisting of social media, internet web documents and high-quality open-source datasets. In our selection of data sources, we exclude certain open-source datasets, such as OSCAR~\cite{OSCAR}, that can contain harmful information.

    \item Curated general data covers a wide range of categories including books (e.g., textbooks, novels), codes, encyclopedias, forums, academic papers, authoritative news, laws and regulations.
    
    
    \item Domain-specific data encompasses popular fields such as finance, taxation, media and publicity, public opinion, and traditional Chinese medicine.
\end{itemize}


\subsubsection{Preprocessing}
We establish a comprehensive data processing pipeline to enhance data quality in all aspects. This pipeline comprises four modules: normalizing, heuristic cleaning, multi-level deduplication, and toxicity filtering. Through comprehensive performance optimization, the pipeline significantly reduces the response time for processing terabyte-scale data to a few hours. Figure~\ref{fig:data_process} illustrates the complete pre-training data processing pipeline. 240 terabytes of raw data are collected for pre-training, and only 10.6 terabytes of high-quality data remain after preprocessing.



\begin{table*}[]
\small
\centering
\setlength{\tabcolsep}{1.1mm}{
\begin{tabular}{@{}cccccc@{}}
\toprule 
{}                                                  &  \textbf{YAYI 2} &  \textbf{ChineseAlpaca 2} &  \textbf{ChatGLM} &  \textbf{Baichuan 1} &  \textbf{XVERSE} \\ \midrule
\textbf{Vocab Size}                                                               & 81920                                    & 55296                                           & 64794                                   & 64000                                      & 100534                                 \\ \midrule
\textbf{Chinese-English bilingual} & {\textbf{ 0.480 $\pm$ {\fontsize{7}{12}\selectfont 0.209}}}     & { 0.502 $\pm$  {\fontsize{7}{12}\selectfont 0.230}}            & { 0.482 $\pm$  {\fontsize{7}{12}\selectfont 0.227}}    & { 0.502 $\pm$  {\fontsize{7}{12}\selectfont 0.239}}       & { 0.640 $\pm$  {\fontsize{7}{12}\selectfont 0.278}}   \\
{ \textbf{Multilingual}}                                      & {\textbf{ 0.476  $\pm$ {\fontsize{7}{12}\selectfont 0.215}}}     & { 0.642 $\pm$  {\fontsize{7}{12}\selectfont 0.434}}            & { 0.551 $\pm$  {\fontsize{7}{12}\selectfont 0.294}}    & { 0.570 $\pm$  {\fontsize{7}{12}\selectfont 0.288}}       & { 0.669 $\pm$  {\fontsize{7}{12}\selectfont 0.286}}   \\ \bottomrule
\end{tabular}}
\caption{Comparison of compression ratio.}
\label{tab:tokenizer_compression_ratio}
\end{table*}

\paragraph{Normalizing}

Through normalization, all raw data are formatted as JSON with keys such as data source, identifier, and content. Additionally, a language detector model is employed for language detection.


\paragraph{Heuristic Cleaning}

We introduce a heuristic multi-level cleaning strategy, building a collaborative filtering mechanism based on chapters, lines, words, and characters. For dozens of data categories such as encyclopedias, Q\&A, news, books, and codes, we devise over a thousand heuristic cleaning rules, tackling issues in formats, contents and encoding. At the chapter level and line level, the strategy concentrates on semantic issues such as garbled characters, logical confusion, and low-quality lines.
At the word level, the emphasis is on eliminating advertising trigger words, while at the character level, the strategy scrutinizes cases of redundant and missing characters.

\paragraph{Multi-level Deduplication}

To filter various duplication patterns, we adopt a multi-level collaborative deduplication strategy, including the chapter-level deduplication based on URL and simHash, the paragraph-level deduplication based on cosine similarity, and the sentence-level deduplication based on prefix-suffix matching.

\paragraph{Toxicity Filtering}


The Internet contains a substantial amount of harmful and false information, including but not limited to pornography, violence, prejudice, discriminatory speech, personal attacks, and illegal activities. To alleviate this problem, we propose a dual-filtering mechanism, which uses a Yayi 2 Checker model based on sensitive words for screening at the first stage and employs a classification model based on quantum heuristic language to complete secondary filtering.

\subsection{Tokenization}

In the international landscape, most LLMs are centered around English, limiting their generalization ability in other languages. Similarly, LLMs released in China tend to focus on bilingual scenarios (Chinese and English), lacking a multilingual training corpus. To enhance the model's comprehension and analytical capabilities across various languages, the YAYI 2 models employ a well-trained multilingual tokenizer.

\paragraph{Training Data}

The tokenizer of YAYI 2 is trained on a 500GB high-quality multilingual corpus, which covers over ten commonly used languages including Chinese, English, French, Russian, etc. The diverse sources of training data encompass web pages, social media, books, newspapers, academic papers, etc.

\paragraph{Vocab Size} 

To support minor languages while maintaining the proficiency in Chinese and English, the YAYI 2 tokenizer expands the vocabulary size to 80,000. Moreover, to harness the tensor parallelization technology and tensor cores efficiently, the vocabulary size needs to be divisible by 128. Thus, we adopt 81,920 as the final vocab size.

\paragraph{Normalization}
 
The YAYI 2 tokenizer adopts a unique approach by directly utilizing raw text for training without undergoing normalization. This strategy ensures the model's adeptness in handling general scenarios.

\paragraph{Algorithm}

By training using the Byte-Pair Encoding (BPE) algorithm~\cite{shibata1999BPE:BytePairencoding} from the Sentence-Piece library~\cite{kudo2018sentencepiece}, the YAYI 2 tokenizer exhibits a robust approach. During training, each digit of a number is intelligently split to facilitate mathematical reasoning. The manually curated vocabulary includes an array of HTML identifiers, common punctuation to enhance segmentation accuracy, and 200 reserved slots for potential applications like adding identifiers during SFT. As a byte-level segmentation algorithm, the YAYI 2 tokenizer excels in handling unknown characters.

\paragraph{Evaluation}
The performance of the tokenizer is measured by the compression ratio, which is defined as follows:
\begin{equation}
     r=\frac{L_{token}}{L_{origin}}
\end{equation}
where $r$ denotes the compression ratio, $L_{token}$ and $L_{origin}$ denote the lengths of the tokenized text and original text, respectively. The lower compression ratio signifies a higher efficiency performance of the tokenizer.


To comprehensively evaluate the YAYI 2 tokenizer’s multilingual performance, we sample data from the SlimPajama~\cite{shen2023slimpajama} dataset and internal data with a length of 10,000 tokens for each, covering Chinese, English, and various minor languages. The results presented in Table~\ref{tab:tokenizer_compression_ratio} reveal that, in both bilingual (CH-EN) and multilingual scenarios, the YAYI 2 tokenizer outperforms other Chinese models such as Baichuan 1~\cite{baichuan1}, ChatGLM~\cite{zeng2022chatglm}, Chinese Alpaca 2~\cite{cui2023chinesealpaca2}, XVERSE~\cite{xverse2023xverse}, boasting a lower compression ratio indicative of superior training and inference efficiency.

\subsection{Model Architectures}

The YAYI 2 models are based on the Transformer architecture~\cite{vaswani2017attention}, embracing the decoder-only structure and training in the autoregressive manner. This architecture, adopted by most prominent LLMs like GPT~\cite{brown2020language}, BLOOM~\cite{workshop2022bloom}, LLaMA~\cite{touvron2023llama, touvron2023llama2} and Baichuan~\cite{baichuan2}, offers advantages such as efficient computation, lower memory usage, and good generalization.



\subsubsection{Positional Embeddings}


Due to the exceptional extrapolation capability, currently there are two popular position encoding methods leveraged by LLMs, i.e., the Rotary Position Embedding (RoPE)~\cite{su2023roformer}, which generates position codes dynamically for the distance between each pair of elements by learning relative position information, and the Attention with Linear Biases Enables Input Length Extrapolation (ALiBi)~\cite{alibi}, which applies a preset offset matrix to the attention score based on the distance between tokens. We empirically find that RoPE shows better adaptation to the accelerate frameworks like Flashattention 2~\cite{dao2023flashattention2} and xFormers~\cite{xFormers2022}. Thus, we opt for RoPE as the chosen positional encoding method.

\subsubsection{Attention Mechanism}


The YAYI 2 models incorporate a distinctive Multi-Query Attention (MQA)~\cite{Shazeer2019FastTD} mechanism to implement Self-Attention, which involves sharing the $W^K$ and $W^V$ weight matrices among heads and concatenating the results. MQA plays a pivotal role in significantly reducing the size of tensors and lowering memory bandwidth requirements for incremental decoding. To enhance the efficiency of calculating the attentions, we leverage the Flashattention 2~\cite{dao2023flashattention2} framework during training to implement the MQA calculation.



\subsubsection{Activations and Normalizations}


Our model incorporates SwiGLU~\cite{shazeer2020glu} as the activation function due to its superior performance and faster convergence. In terms of the regularization method, we employ RMSNorm~\cite{zhang2019RMS}, which only focuses on the rescaling invariance and performs regularization to the summed inputs simply based on the root mean square. Compared to the commonly used Layer Normalization~\cite{ba2016layernorm}, RMSNorm can approximately reduce computation time by 7\%-64\%.

\subsection{Model Training}

\subsubsection{Computing Cluster}
\label{sec:computing_cluster}


YAYI 2 models are trained on a cluster comprising over 1000 A800 GPU servers. This cluster's nodes are interconnected through an InfiniBand (IB) network, facilitating high-speed direct memory access and data transfer. GPUs within each node are linked through high-bandwidth, low-latency NVLink connections. To optimize cluster management of code, data, and model checkpoints, an SSD hard drive is implemented as the shared storage for the whole cluster using the Network File System (NFS). Addressing common challenges in large-scale cluster management, such as resource allocation, job scheduling, and scalability, we enhance the SLURM (Simple Linux Utility for Resource Management) system for resource management and job scheduling. Additionally, an anomaly alert module is also added to monitor the real-time running status of the cluster in case of hardware failures and unhandled program exceptions.

\subsubsection{Training Strategy}


\paragraph{Distributed Training} To keep a balance between GPU memory utilization and communication efficiency, the Zero Redundancy Optimizer (ZeRO)~\cite{rajbhandari2020zero} stage 3 is applied, which works in conjunction with gradient checkpointing, significantly improving the GPU memory utilization. As expected, the average processing speed of GPUs reaches 600 tokens/s, with tensor core utilization rate of 65\%, showcasing superior performances in large-scale clusters~\cite{touvron2023llama}.



\paragraph{Optimizer} The AdamW~\cite{Loshchilov2017DecoupledWD} is used for training. Unlike Adam~\cite{diederik2015adam}, AdamW achieves higher computational efficiency, superior generalization, and faster convergence speed. For parameters of AdaW, the $\beta_1$ and $\beta_2$ are set be to 0.9 and 0.95, respectively. The weight decay is 0.1. The model training is warmed up with the learning rate from $5\times10^{-5}$ to $1\times10^{-4}$ for the first 2000 steps.

\begin{figure}
  \centering
      \includegraphics[width=0.47\textwidth]{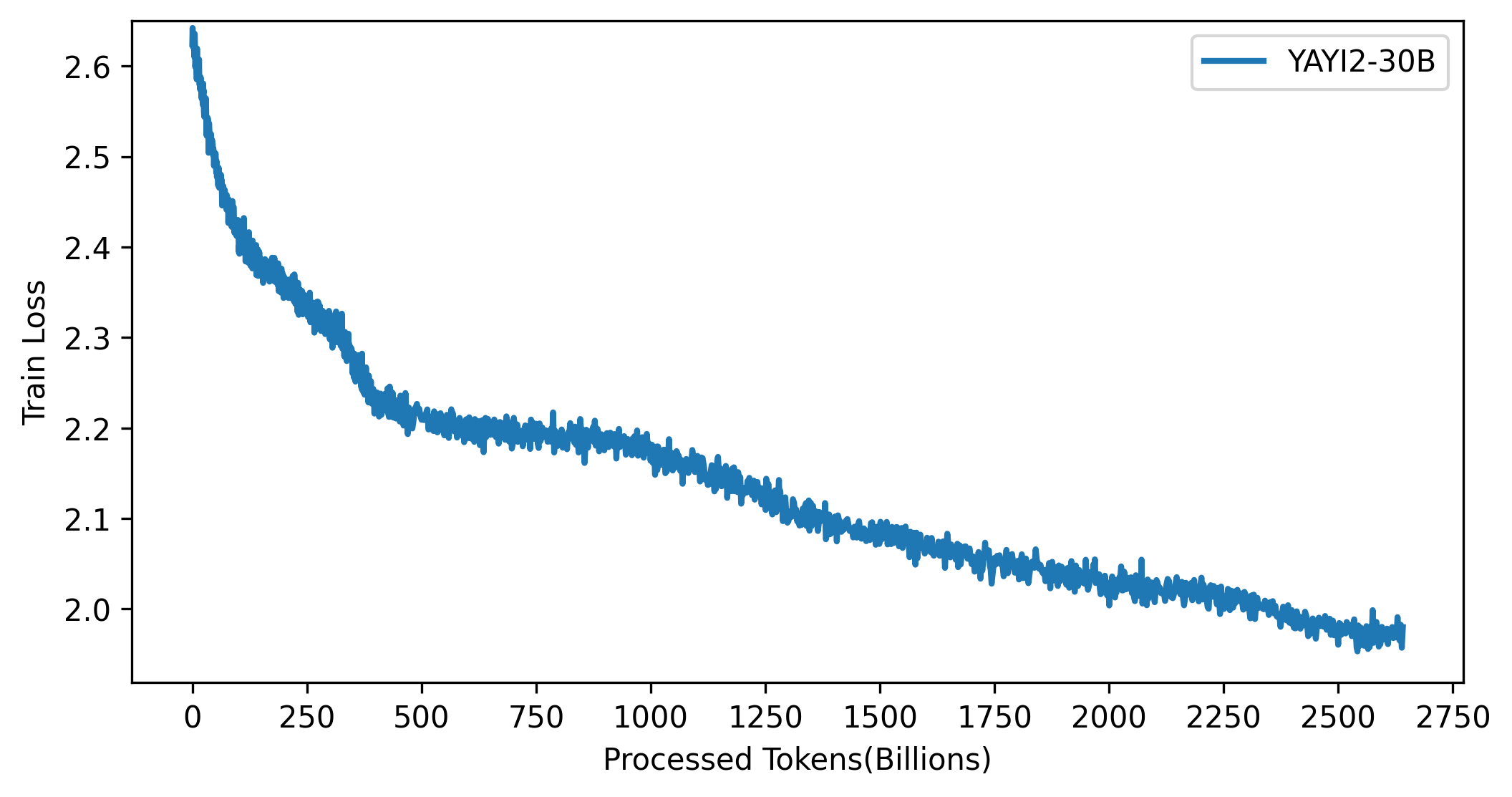}
  \caption{The training loss of YAYI 2-30B.}
  \label{fig:pretrain_loss}
\end{figure}

Figure~\ref{fig:pretrain_loss} shows the final training loss of YAYI2-30B.

\subsubsection{Training Tricks}
\paragraph{Data Pre-allocation}

Maintaining a stable data distribution is pivotal for improving model performances. Large jitter in data distribution can be harmful to model convergence. To precisely control the data distribution, we design a data pre-allocation mechanism based on file indices. This mechanism builds a global file index table and allocates data files to each GPU before training, guaranteeing consistent data distribution across training steps. According to whether the quantity of data is fixed, pre-training data can be divided into static data and dynamic data. The quantity of static data does not change with time, and mainly includes knowledge data such as books, ancient poetry, textbooks, academic papers, encyclopedia knowledge bases, etc. The quantity of static data is limited but of high quality, whereas dynamic data exhibits a vast quantity but with lower quality. The quantity of dynamic data continues to grow over time, mainly including current news data such as web pages, newspapers, social media, etc. To reduce model hallucinations, we upsample static data and downsample dynamic data by increasing and decreasing file indices, respectively.

\paragraph{Lazy Loading}

When loading binary model checkpoints, since each GPU in one node needs to pre-load the model weights from the node's CPU memory into its GPU memory, the CPU memory may overflow under different configurations of computing clusters. By introducing a lazy loading strategy, i.e. allowing different GPUs to start the pre-loading process sequentially, we reduce the peak memory usage during the model loading phase and effectively avoid CPU memory overflow.


\paragraph{Training Restart}


With the expansion of the computing cluster, training tasks are prone to be interrupted due to various software and hardware issues. To minimize idle time of the training cluster and restart training from intermediate checkpoint, we optimize preventive measures for common problems such as GPU crash, disk space exhaustion, deadlocks, etc. Specifically, we perform automated interventions from three aspects: logging, exception alerts, and automatic restarts.
    
\begin{itemize}
    \item \textbf{Logging.} We maintain detailed logs of the current training task status, including model training outputs and data usage states.
    \item \textbf{Exception alerts.} By monitoring GPU utilization and the update timestamp of log files, we establish an auto-alert mechanism via instant messaging. The types of malfunctions is also detected and notified.
    \item \textbf{Automatic restarts.} Based on the type of malfunction, the training cluster adopts corresponding restart strategies. For example, when some GPU crashes, problematic nodes are removed, and standby nodes are incorporated into the training cluster before restarting the training process.
    
\end{itemize}

\begin{table*}[htbp]
\small
\centering
\setlength{\tabcolsep}{8mm}{
\begin{tabular}{@{}llc@{}}

\toprule
\textbf{Task Type} & \textbf{Description}    & \textbf{Weight} \\ \midrule
Text Generation          & Generate articles, outlines, schemes, etc.      & 30\%            \\
Reading Comprehension          & Answer questions based on given context.  & 18\%            \\
Open QA          & Knowledge, common sense, and other questions.	         & 10\%            \\
Creative Inspiration          & Write poetry, design, naming, script creation, etc.   & 10\%            \\
Information Extraction          & Extract content from context, output in a specified format. & 8\%             \\
Chit-chat Role Play          & Daily consultations, chat, and role-playing.     & 5\%             \\
Text Rewriting          & Change style, change language, etc.         & 5\%             \\
Abstraction \& Summarization          & Summarize titles or abstracts, etc.         & 4\%             \\
Text Classification          & Text classification task.           & 3\%             \\
Text Translation          & Multilingual translation tasks.          & 2\%             \\
Code Capability          & Code generation, completion, commenting, etc.  & 2\%             \\
Logical Reasoning          & Mathematical and reasoning tasks.       & 2\%             \\
Other Tasks          & Tasks not classified into the above categories.   & 1\%           \\
\bottomrule
\end{tabular}}
\caption{General tasks and descriptions.}
\label{tab:multi-task&description}
\end{table*}

\section{Alignment}

The alignment process for YAYI2-30B consists of two crucial stages: Supervised Fine-Tuning (SFT) and Reinforcement Learning from Human Feedback (RLHF).
\subsection{Supervised Fine-Tuning}

\subsubsection{Instruction Dataset}





Instruction data for YAYI encompasses manually annotated high-quality instruction data and open-source SFT datasets. We strictly review the instruction data in terms of format and content. For data format, we check for missing line breaks. For the content, we: (1) examine the completeness of content (i.e., truncated answers, incomplete information, and repeated generations); (2) ensure the consistency of language in instructions and answers, except for translation tasks; (3) confirm that the generated answers follow the given instructions; (4) ensure that the generated responses are free from hallucination; (5) verify that the answers comply with laws and regulations; (6) scrutinize the human values in the instruction-answer pairs.


For the data format, content completeness, and language consistency, a classification model is trained to evaluate the open-source instruction data and the auto-generated data. For the instruction compliance and the hallucination issue, we systematically inspect data in batches through manual verification. Data sources within each batch are consistent. A batch of data is dismissed if it displays poor compliance with instructions or has many hallucination issues. For safety concerns, see Section~\ref{sec:safety_fine-tuning}.




After filtering and review, we identify high-quality data to ensure balanced sampling for training. To promise the data diversity for SFT, we sample data across different task types, language categories, context lengths, and data sources, where the distribution of general task types is outlined in Table~\ref{tab:multi-task&description}.




Following OpenAI's Chat Markup Language (ChatML) format, the SFT for YAYI adheres to a structured multi-turn conversation involving three roles: system, human, and Yayi. The system defines global role information and identity settings, the human represents the user, and YAYI symbolizes the large model. Identifiers for these roles are denoted as "<system>", "<human>", and "<yayi>" for clarity.



\subsubsection{Training Details}


Aligned with the pre-training stage, the YAYI 2 models employ a distributed training framework for SFT. The training utilizes BF16 floating-point precision to enhance efficiency and employs the AdamW optimizer with $\beta_1$ set as 0.9, $\beta_2$ set as 0.96, and $\epsilon$ set as 1e-8. The learning rate warm-up steps constitute approximately 10\% of the total steps, gradually increasing to a peak learning rate of 5e-6. To prevent overfitting, the weight decay is set as 1e-3.

To accommodate various instruction lengths during training, including short, long, single-turn conversation, and multi-turn conversation instructions, we progressively adjust the context window from 2,048 to 4,096 and ultimately 8,192. The computing cluster is the same as Section~\ref{sec:computing_cluster}.


\begin{figure*}[htb]
    \centering
    \includegraphics[width=0.85\textwidth, trim=0 10 0 30]{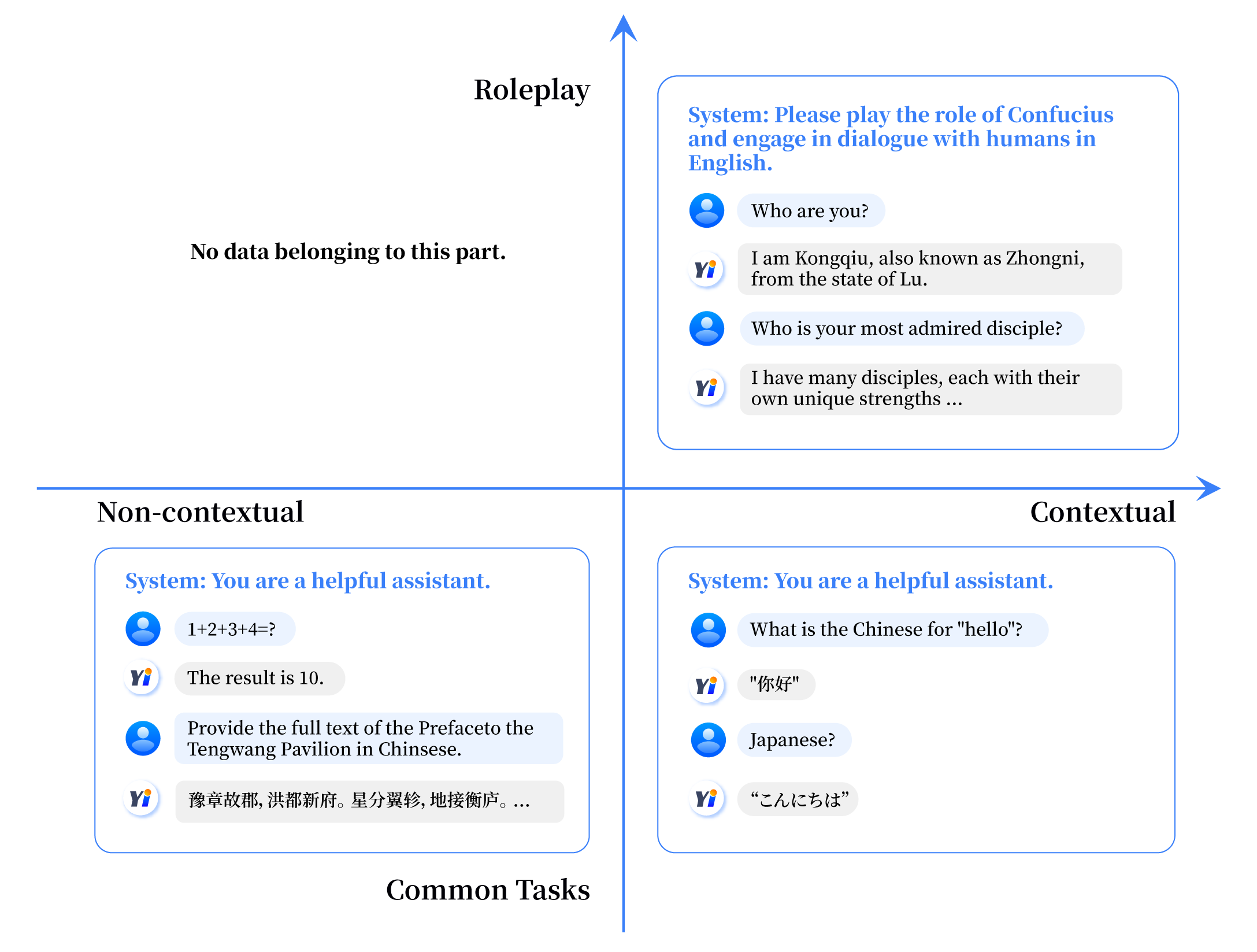}
    \caption{Dimensions of multi-turn conversation.}
    \label{fig:multi-turn-data}
\end{figure*}

\subsubsection{Long Instructions}
To bolster the model's capability in handling lengthy texts, a batch of long SFT data is built, encompassing both the long input type and long output type. The long input data includes tasks like long text summarization, reading comprehension, and other complex instructions. The long output data involves generating long articles, multi-level outlines, and research reports, etc.


\subsubsection{Multi-Turn Conversation}
We build multi-turn conversation data from two dimensions:
\begin{itemize}
    \item Context Relevance Dimension: including context-relevant and context-irrelevant multi-turn conversation data. Context-relevant conversations involve human questions related to the previous content, while context-irrelevant conversations comprise questions unrelated to the ongoing conversation.
    \item Role Information Dimension: including multi-turn conversations for general tasks (without role information) and role-played multi-turn conversations.
\end{itemize}


In the course of instruction data generation, we applied a nuanced approach, segmenting the data into distinct role information dimensions. This segmentation encompassed multi-turn conversations tailored for general tasks and role-played multi-turn conversations, strategically intertwined with contextual relevance for data synthesis. In the realm of general tasks, multi-turn conversations featured instances both relevant and irrelevant to the ongoing context. In contrast, role-played multi-turn conversations, distinguished by their succinct and context-driven nature, exclusively factored in context-relevant scenarios. This conceptualization is succinctly depicted in Figure~\ref{fig:multi-turn-data}.

\begin{itemize}
    \item Context-relevant multi-turn conversation data in general tasks: In this regard, we devise a meticulous algorithm for data generation. Commencing with a randomly sampled human question data-seeding from the instruction database, the model generates an initial response. Subsequently, leveraging the extant context, we systematically formulate related questions and amalgamate the contextual content to prompt the model for generating successive rounds of responses. This iterative process results in the creation of context-relevant multi-turn conversation data tethered to the original data-seeding.
    \item Context-irrelevant multi-turn conversation data in general tasks: In this regard, we independently draw random batches of task-type-similar and task-type-irrelevant single-turn data. Through statistical scrutiny, it emerges that human queries in a single multi-turn conversation exhibit a propensity for thematic similarity. Guided by this observation, we sample and concatenate analogous task-type data or devised mixed-sample data, mirroring scenarios where humans pose multiple queries related to the same task (e.g., prompting the model to generate poetry repeatedly) or varied tasks within a single multi-turn conversation (e.g., initiating poetry generation followed by mathematical problem-solving).
    \item Role-played multi-turn conversations: Prompt generation begins by randomly assigning roles to the YAYI model, encompassing task-centric roles (e.g., traditional Chinese medicine practitioner, lawyer, financial analyst) and specific character roles (e.g., Confucius, Sun Wukong, Zhang Fei). Based on the speaking style and personality traits inherent in these roles, we simulate multi-turn conversations involving human participants. Following rigorous quality assessments, it assumes the role of the model's multi-turn conversation training dataset.
\end{itemize}

\begin{figure}
  \centering
      \includegraphics[width=0.45\textwidth, trim=0 50 0 0]{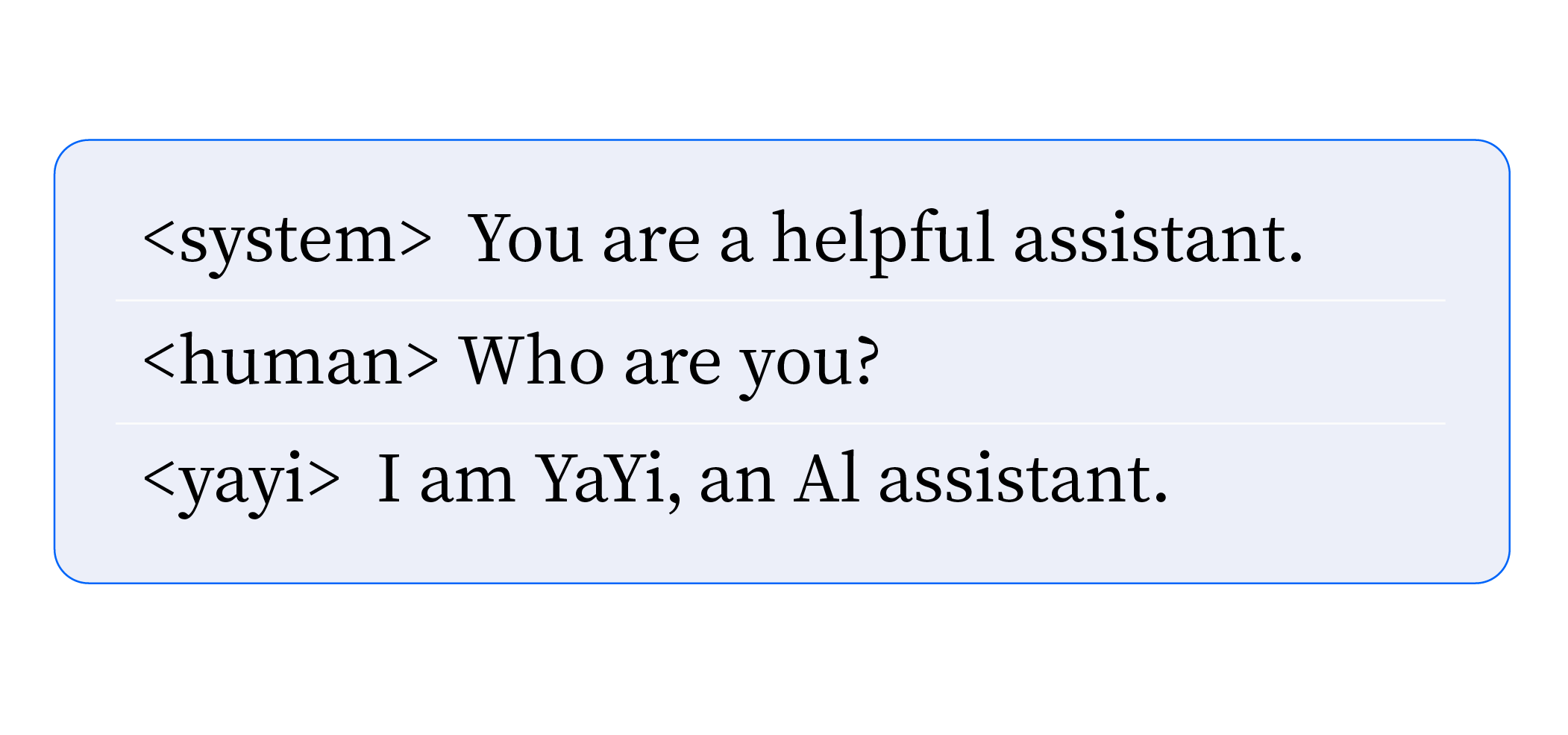}
  \caption{Multi-turn conversation data format.}
  \label{fig:multi_format}
\end{figure}

\begin{figure}
  \centering
      \includegraphics[width=0.45\textwidth, trim=0 50 0 0]{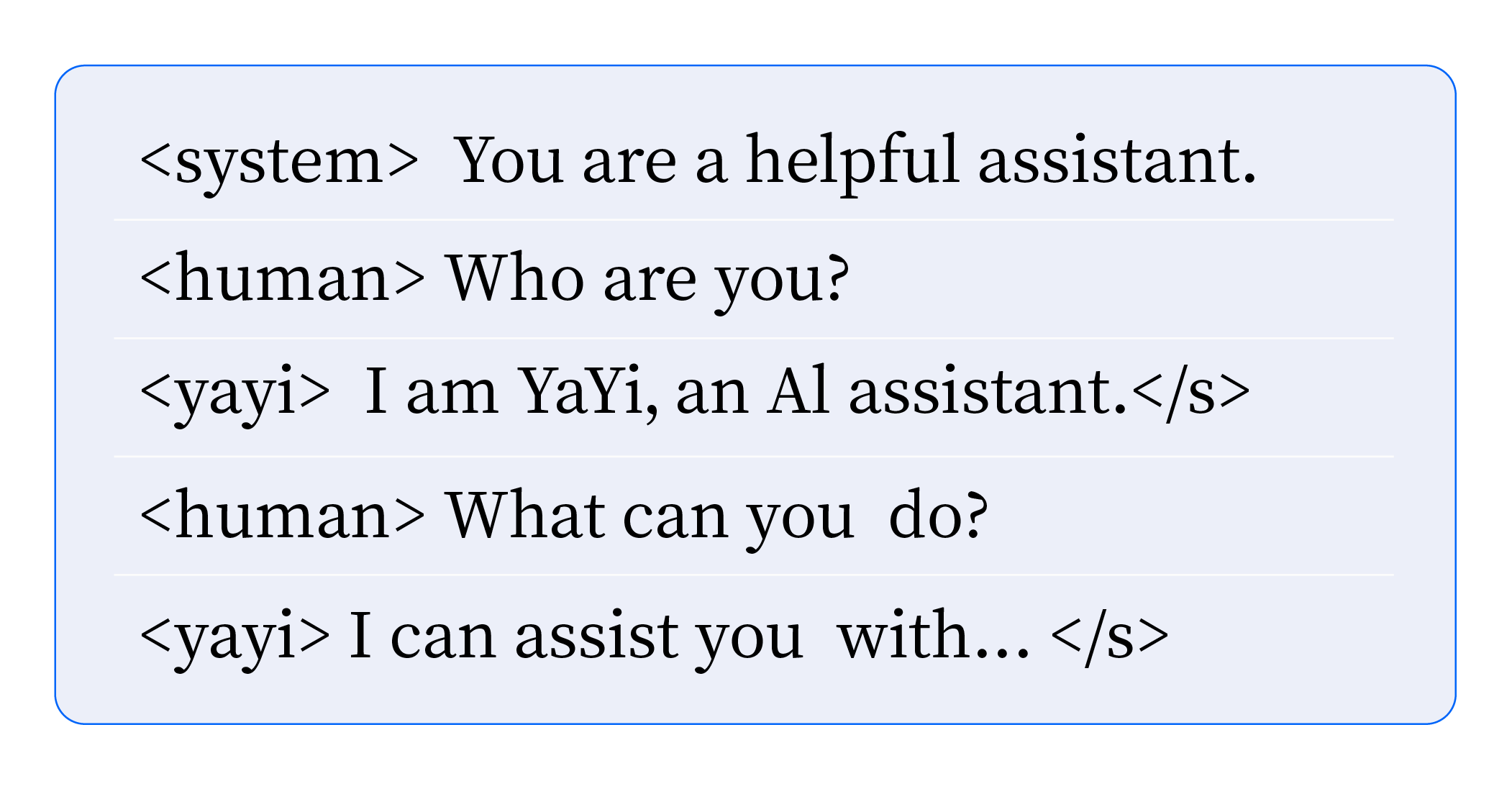}
  \caption{Roleplay data format.}
  \label{fig:roleplay_format}
\end{figure}

The format of the multi-turn conversation training data aligns with that of the single-turn instruction task training. It commences with globally defined role information, alternates between the user and the YAYI model, and concludes each YAYI model's reply with an end-of-sequence token "</s>." The format is illustrated below.



\begin{figure}
    \centering
    \includegraphics[width=0.54\textwidth,trim=100 0 0 0,clip]{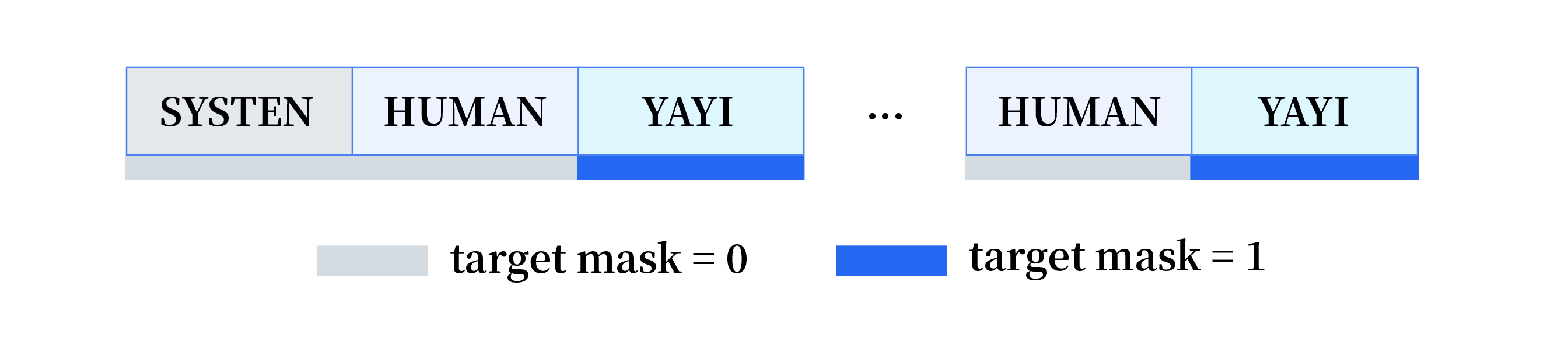}
    \caption{Computation of multi-turn conversation loss.}
    \label{fig:multiround_loss}
\end{figure}

During model training, we only compute the loss for the output of each turn in multi-turn conversations, as depicted in Figure~\ref{fig:multiround_loss}. This strategic approach steers the model toward generating high-quality conversation content, circumventing unnecessary calculations for irrelevant content. This targeted methodology significantly augments training efficiency.

\subsubsection{Domain Tasks}
The YAYI large model is meticulously tailored to real-world business scenarios, encapsulating a diverse spectrum of hundreds of tasks spanning finance, media, law, healthcare, and beyond. Through manual annotation and review, we construct a series of domain-specific data for SFT, aiming to hone the model's prowess in navigating authentic business challenges. 



\subsection{Reinforcement Learning from Human Feedback}
Despite the commendable performances of supervised fine-tuning across various tasks, the efficacy of the proposed model is contingent on the quality of annotated data and is susceptible to overfitting. To overcome these limitations and further elevate the YAYI models' capacity for generalization, we turn to reinforcement learning from human feedback \cite{ouyang2022training}. This methodology aims to align the models' generated content more closely with human preferences. Specifically, a reward model is trained to predict human preferences, and the Proximal Policy Optimization (PPO) \cite{schulman2017PPO} algorithm is employed to reinforce the YAYI model, guiding it toward generating responses that better resonate with human expectations.

\subsubsection{Reward Model}
To collect high-quality and well-balanced instructions, a meticulous instruction sampling strategy is implemented. Initially, a semantic deduplication system is utilized for a vast instruction set. Subsequently, a two-tier task subdivision is employed with a dynamic weighted sampling strategy to maintain instructional equilibrium within each category.

Given a prompt, the YAYI 2 chat model generates two responses, employing distinct sampling strategies. Expert annotators evaluate these responses across four dimensions: format correctness, content completeness, content accuracy, and instruction compliance. These evaluations are employed for the continuous refinement of the reward model's performance.

The reward model is trained starting with the YAYI chat model after SFT. Notably, for performance stability, a reward token is appended to each data point. The embedding features of this token are then utilized to predict the ultimate reward. Throughout training, the reward model exhibits an escalating trend in discriminative accuracy as the quality gap between the two responses widens.



\subsubsection{Reinforcement Learning via PPO}

The PPO algorithm is adopted for reinforcement learning, encompassing four models: the policy model (responsible for response generation, requiring optimization), the reference model (providing a fixed-parameter reference for the policy model), the reward model (assessing response quality with fixed parameters), and the value model (learning token values and requiring optimization). The value model undergoes a warm-up stage of 50 training steps during the training process. Both the value model and policy model are updated using the standard PPO algorithm. To maintain training stability, clipping and normalization techniques are also applied.

\section{Inference}

\subsection{Long-Context Reasoning}


The YAYI 2 models have significantly enhanced their capacity for processing lengthy texts and multi-turn conversations by leveraging an extended context window. While mainstream proprietary models, like GPT-4-Turbo, have extended their context length to 128K, open-source LLMs, such as Llama, typically support a context length of 4K. In this technical report, we augment the YAYI 2 models' ability to handle extensive contextual information by extending its extrapolation capabilities based on scalable features of the RoPE position embedding method. 

Current research in RoPE extrapolation primarily explores two directions: methods based on sliding windows and methods based on adjusting rotation angles. Given the loss of global low-frequency information associated with sliding windows, recent studies concentrate more on adjusting the encoding rotation angle. The YAYI 2 models adopt the YaRN method~\cite{peng2023yarn} for RoPE extrapolation, integrating NTK with sliding window methods to mitigate the collapses in ultra-long contexts.

\begin{figure}
  \centering
      \includegraphics[width=0.5\textwidth, trim=0 0 0 0]{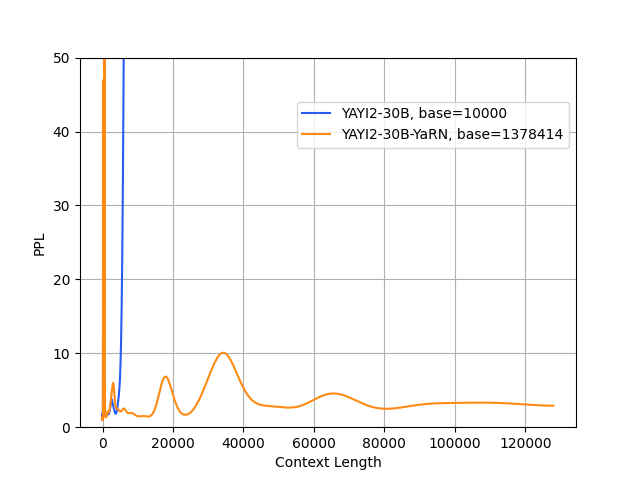}
  \caption{Perplexity of different configurations for extrapolation.}
  \label{fig:extrapolation_128k}
\end{figure}

Figure~\ref{fig:extrapolation_128k} shows that YAYI2-30B with YaRN has significantly lower perplexity and is more stable, which demonstrates that the effectiveness of NTK with sliding window for extrapolation.

\subsection{Diverse Hardware Inference Adaptation}



In addition to NVIDIA GPUs, the YAYI 2 models have been adapted for efficient inference on the Huawei Ascend 910B hardware. To address the challenges posed by the large parameter count for a 30B model during single-GPU inference, we adopt a distributed inference solution, which involves using multi-GPUs for inference. This process entails compiling the target strategy network to obtain the distributed strategy file. Thus based on the splitting strategy, the model weights are partitioned and loaded onto each GPU for the following inference procedure.



\begin{figure*}[h]
  \centering
      \includegraphics[width=0.7\textwidth,trim= 0 0 -220 0]{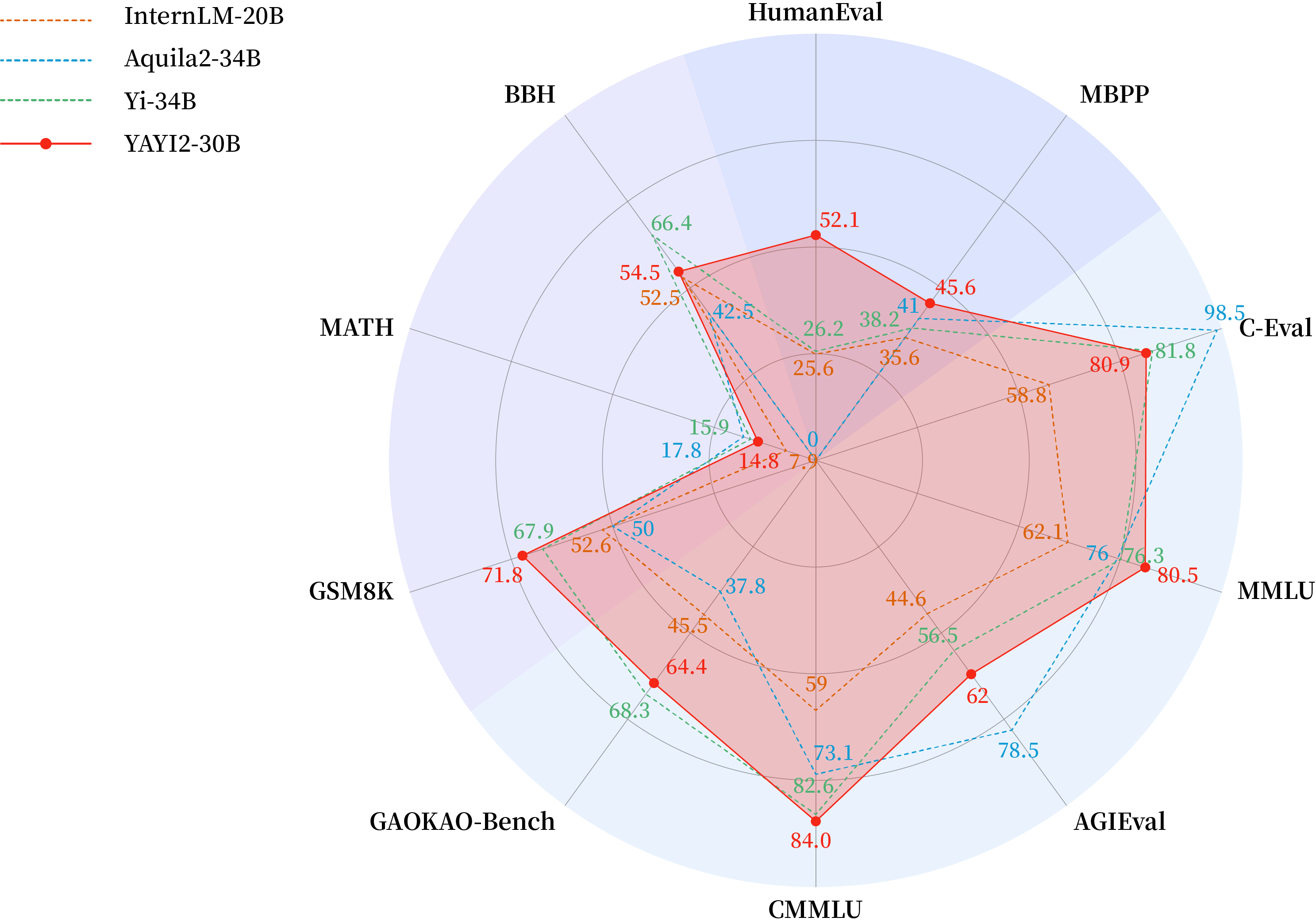}
  \caption{Results of similar sized LLMs on 10 benchmarks.}
  \label{fig:spider_comparison}
\end{figure*}

\section{Safety}

\subsection{Pre-training Stage}
The pre-training data-preparation phase prioritizes and strictly adheres to data security protocols to ensure the integrity and compliance of the data. A comprehensive strategy is deployed, incorporating a robust data security filtering and classification system.

This system's primary objective is to identify and exclude potentially harmful and security-related content, preventing the model from learning and reproducing inappropriate information. The specific categories of safety-related information include:

\begin{itemize}
    \item Sensitive information. Confidential internal corporate data, such as undisclosed financial reports and research materials, is filtered out to prevent intellectual property infringement issues. Other sensitive information includes personal privacy data, including but not limited to personal identity information, contact details, bank accounts, and medical records.
    
    \item Inappropriate content. Inappropriate content includes hate speech, violence, discrimination (e.g. ethnic and gender discrimination), extremist speech, pornography and other indecent content.

    \item Content Violating Laws and Regulations. Copyrighted materials are removed to ensure that protected works under copyright are not used illegally or included in training data. 

    \item Misleading and Inaccurate Information. Misinformation includes fake news, rumors, and other potentially misleading content. Anti-scientific, pseudoscientific, and inaccurate medical health information are also removed to curb the spread of misinformation.
\end{itemize}

These strategies are implemented during the data source selection and data processing steps. Initially, data source selection involves strict adherence to reputable channels to sidestep intellectual property disputes. In the preliminary screening of text data, a Deterministic Finite Automaton (DFA)-based sensitive word filtering mechanism is applied. For Chinese text, the segmentation system is expanded, incorporating a specialized segmentation library to enhance filtering accuracy.

Furthermore, an efficient text classification model is developed and trained to identify and eliminate text containing inappropriate content. The training set covers categories such as pornography, violence, discrimination, hate speech, personal safety infringement, and illegal content. To broaden the model's recognition capabilities, sample translation between Chinese and English is conducted, diversifying training samples. Medical professional data is specifically included in training samples to prevent medical-related text misclassification. These above two critical steps significantly enhance the security of the training data and lay a robust foundation for subsequent model training.

\begin{table*}[]
    \centering
    \resizebox{0.85\textwidth}{!}{
    \begin{tabular}{lccccc} 
        \toprule
        \multirow{2}{*}{\textbf{Model}}&\textbf{C-Eval(val)}&\textbf{MMLU}&\textbf{AGIEval}&\textbf{CMMLU}&\textbf{GAOKAO-Bench}\\
        &5-shot&5-shot&	3/0-shot&	5-shot&	0-shot\\
        \midrule
        MPT-30B&	--&	46.9&	33.8&--	&--	\\
        Falcon-40B&	--&55.4&37.0&--&--\\
        LLaMA2-34B&	--&62.6&	43.4&--	&--	\\
        Baichuan2-13B&	59.0&	59.5&	37.4&	61.3&	45.6\\
        Qwen-14B&	71.7&	67.9&51.9	&	70.2&	62.5\\
        InternLM-20B&	58.8&	62.1&	44.6&	59.0&	45.5\\
        Aquila2-34B&	\textBF{98.5}&	76.0&43.8	&	78.5&37.8	\\
        Yi-34B&	81.8&	76.3&	56.5&	82.6&\underline{68.3}\\
        Qwen-72B&	\underline{83.3}&	\underline{77.3}&\underline{61.8}	&	\underline{83.6}&	\textBF{86.0}\\
        \midrule
        YAYI2-30B&80.9&\textBF{80.5}&\textBF{62.0}&\textBF{84.0}&64.4\\
        \bottomrule
    \end{tabular}
    }
    \caption{Evaluation results on knowledge and language understanding. The best results are in \textbf{bold} and the second are \underline{underlined}.}
    \label{tab:knowledge}
\end{table*}

\subsection{Fine-tuning Stage}
\label{sec:safety_fine-tuning}

A safety auditing algorithm combining regular matching with machine learning models is designed for safety classification in the fine-tuning stage. Among the SFT data, safety instruction data is categorized into positive guidance and refusal-to-answer:
\begin{itemize}
    \item Positive guidance category: Questions containing statements misaligned with human values or contradicting objective facts require the model to correct inappropriate content and provide a positive guidance response in line with human values.
    \item Refusal-to-answer category: Questions involving illegal issues or those contravening relevant policies and laws prompt the model to express apologies, inform users of inappropriate content, and refuse to provide an answer.
\end{itemize}

Joint training of safety-enhanced instruction data with general tasks and domain tasks is carried out to prevent catastrophic forgetting and enhance the model's security.

\section{Evaluations}

\subsection{Baseline Models}


In this section, we evaluate the YAYI 2 base model's performance against a series of open-source models with similar parameter sizes on standard benchmark datasets. The evaluation dimensions encompass knowledge and language understanding, mathematical reasoning, and programming ability. Comparative base models include MPT-30B \citep{mn2023introducing}, Falcon-40B \citep{almazrouei2023falcon}, LLaMA 2-34B \citep{touvron2023llama2}, Baichuan 2-13B \citep{baichuan2}, Qwen-14B\&72B \citep{bai2023qwen}, InternLM-20B \citep{internlm2023internlm}, Aquila 2-34B \citep{baai2023Aquila2} and Yi-34B \citep{01ai2023yi}.

\subsection{Evaluation Results}



We use accuracy as the primary metric and, if available, report the results of comparative models evaluated by OpenCompass \cite{2023opencompass}, taken from the leaderboard of the OpenCompass official website\footnote{\url{https://opencompass.org.cn/leaderboard-llm}, evaluation results reference date: Dec. 15, 2023.}. The reported results of YAYI2-30B model are also evaluated by the source code at the OpenCompass Github repo. For the models that have not been evaluated by the OpenCompass, including MPT-30B, Falcon-40B and LLaMA 2-34B, we use the results reported by \citet{touvron2023llama2}. Note that on some benchmarks there can be some slight differences between the evaluations conducted by OpenCompass and \citet{touvron2023llama2}. See Figure~\ref{fig:spider_comparison} for the overall comparison with three similar sized LLMs, including InternLM-20B, Aquila2-34B and Yi-34B.

\subsubsection{Knowledge and Language Understanding}


The evaluations regarding knowledge cover various benchmarks, including MMLU \citep{hendrycks2020measuring}, C-Eval validation set \citep{huang2023c}, CMMLU \citep{li2023cmmlu}, AGIEval \citep{zhong2023agieval} and GAOKAO-Bench \citep{zhang2023evaluating}.


\begin{itemize}
    \item MMLU: English interdisciplinary knowledge evaluation benchmark, covering multiple choice questions from 57 disciplines in STEM, humanities, social sciences, and other fields.
    \item C-Eval: Chinese comprehensive exam evaluation benchmark, consisting of 13,948 multiple choice questions, with four different levels of difficulty, covering knowledge across 52 disciplines.
    \item AGIEval: Benchmark for knowledge reasoning ability in both Chinese and English, including questions in various fields such as SAT, college entrance examination, and judicial exam.
    \item CMMLU: Chinese benchmark assessing knowledge reasoning ability, including 67 single-choice questions across various themes in natural science, humanities and social sciences, and everyday life.
    \item GAOKAO-Bench: Chinese benchmark for knowledge reasoning ability, including major questions from national college entrance exams from 2010 to 2022, from which objective questions are selected to evaluate the model.
\end{itemize}

We report the 3-shot (for MPT-30B, Falcon-40B and LLaMA 2-34B) or zero-shot (for other models) evaluation results on AGIEval and GAOKAO-Bench, and 5-shot results on MMLU, C-Eval and CMMLU. Table~\ref{tab:knowledge} shows the detailed results of our proposed model in the comparative experiments on these benchmarks. Our model outperforms other models on MMLU, AGIEval and CMMLU benchmarks, even surpassing the Qwen-72B with a much larger parameter size.

\begin{table} 
    \centering
    \resizebox{\columnwidth}{!}{
    \begin{tabular}{lccc} 
        \toprule
        \multirow{2}{*}{\textbf{Model}}&\textbf{GSM8K}&\textbf{MATH}&\textbf{BBH}\\
        &8/4-shot&	4-shot&	3-shot\\
        \midrule
        MPT-30B&15.2&3.1&38.0 \\
        Falcon-40B& 19.6& 5.5&	37.1\\
        LLaMA2-34B&	42.2& 6.2&	44.1\\
        Baichuan2-13B&	52.6&	10.1&	49.0\\
        Qwen-14B&	61.6&	\underline{25.2}&	53.7\\
        InternLM-20B&	52.6&	7.9&	52.5\\
        Aquila2-34B&	50.0&	17.8&	42.5\\
        Yi-34B&	67.9&	15.9&	\textBF{66.4}\\
        Qwen-72B&	\textBF{77.6}&	\textBF{35.1}&	\underline{63.7}\\
        \midrule
        YAYI2-30B&\underline{71.2}&14.8&54.5\\
        \bottomrule
    \end{tabular}
    }
    \caption{Evaluation results on mathematical reasoning.}
    \label{tab:math}
\end{table}

\subsubsection{Math and Logic Reasoning}


In the domain of mathematics and reasoning, our model is evaluated on three prominent benchmarks: GSM8K \citep{cobbe2021training}, MATH \citep{hendrycks2021measuring} and BBH \citep{suzgun2022challenging}. We use accuracy as the principal evaluation metric.


\begin{itemize}
    \item GSM8K: A benchmark dataset designed for mathematical reasoning, encompassing 1,319 elementary math word questions.
    \item MATH: Comprising 5,000 challenging mathematical questions spanning diverse domains such as linear algebra, geometry, and probability.
    \item BBH: A subset of the BIG-Bench dataset, featuring 23 highly challenging tasks encompassing logic reasoning, common-sense understanding, and mathematics. Its objective is to challenge the model with more intricate reasoning and language-understanding tasks.
\end{itemize}



We report the 8-shot (for MPT-30B, Falcon-40B and LLaMA 2-34B) or 4-shot (for other models) evaluation results on GSM8K, 4-shot results on MATH, and 3-shot results on BBH. Upon examination of Table~\ref{tab:math}, the YAYI 2 base model has achieved the best performance on the GSM8K benchmark among models with comparable parameter sizes.

\begin{table} 
    \centering
    \resizebox{0.89\columnwidth}{!}{
    \begin{tabular}{lcc} 
        \toprule
        \multirow{2}{*}{\textbf{Model}}&\textbf{HumanEval}&\textbf{MBPP}\\
        &0-shot&  3-shot\\
        \midrule
        MPT-30B&25.0&32.8 \\
        Falcon-40B&0.6&29.8\\
        LLaMA2-34B& 22.6 &33.0\\
        Baichuan2-13B&  17.1& 30.8\\
        Qwen-14B& 32.3& 39.8\\
        InternLM-20B& 25.6& 35.6\\
        Aquila2-34B&  0.0&  41.0\\
        Yi-34B& 26.2& 38.2\\
        Qwen-72B& \underline{33.5}& \textBF{51.6}\\
        \midrule
        YAYI2-30B&\textBF{53.1}&\underline{45.8}\\          
        \bottomrule
    \end{tabular}
    }
    \caption{Evaluation results on programming.}
    \label{tab:code}
\end{table}

\subsubsection{Programming}


In the evaluation of programming capabilities, the evaluation benchmarks include HumanEval \citep{chen2021evaluating} and MBPP \citep{austin2021program}


\begin{itemize}
    \item HumanEval: A dataset comprising 164 programming questions, each featuring a function signature, documentation string, subject code, and an average of 7.7 unit tests. Covering aspects of language understanding, reasoning, algorithms, and basic mathematics, it serves as a comprehensive assessment of the model's proficiency in code generation.
    \item MBPP: A coding benchmark consisting of 500 beginner-level Python programming questions.
\end{itemize}


The primary evaluation metric is pass@1, indicating the model's success rate in generating the correct code on the first attempt. 



Following the evaluation method of OpenCompass, we report the zero-shot results on HumanEval and 3-shot results on MBPP. Table~\ref{tab:code} demonstrates our model's standing as the pinnacle performer among models with comparable parameter sizes, and even significant superiority over the much larger Qwen-72B on the HumanEval benchmark. In summary, our model showcases remarkable competence across knowledge understanding, mathematical reasoning, and programming benchmarks, validating the effectiveness of our model.


\section{Conclusions}

In this technical report, we propose the multilingual YAYI2-30B LLMs with a specific focus on Chinese-related applications. We introduce the distributions of the pre-training dataset, as well as the preprocessing pipeline. The YAYI2-30B models follow the popular decoder-only model architecture, and adopt FlashAttention 2 and MQA to speed up training and inference. We also reveal the pre-training details, including computing clusters, training strategies and tricks, which we believe will greatly benefit the industry practitioners. We further show how to build the instruction dataset for instruction tuning, and the YAYI 2 models' support for long instructions, multi-turn conversations and domain-specific applications. The RLHF process is further applied to better align with human values and ensure safety. The YAYI 2  base model is evaluated on three types of benchmarks, including knowledge and Language understanding, math and logic reasoning, and programming. Extensive experimental results show that the proposed model achieves superior performances over similar-sized open-source LLMs on multiple benchmarks, including MMLU, AGIEval, CMMLU, GSM8K, HumanEval and MBPP. Especially on the MMLU, AGIEval, CMMLU and HumanEval benchmarks, our model can even outperform the larger-sized Qwen-72B with considerable margins.

Although we have adopted various methods to ensure safety and reduce hallucinations, the YAYI 2 models still can produce harmful content or fabricate "facts", so the model users are highly encouraged to review the answers, especially in safety-critical situations. The model users are also advised to prevent the misuse of the YAYI 2 models and abide by related laws and regulations. The YAYI 2 models are still under active development, and all suggestions and feedback are welcomed.

\bibliography{custom}

\begin{thebibliography}{51}
\expandafter\ifx\csname natexlab\endcsname\relax\def\natexlab#1{#1}\fi

\bibitem[{01-AI(2023)}]{01ai2023yi}
01-AI. 2023.
\newblock Yi: {A} series of large language models trained from scratch by
  developers at 01-ai.
\newblock \url{https://github.com/01-ai/Yi}.

\bibitem[{Almazrouei et~al.(2023)Almazrouei, Alobeidli, Alshamsi, Cappelli,
  Cojocaru, Debbah, Goffinet, Heslow, Launay, Malartic, Noune, Pannier, and
  Penedo}]{almazrouei2023falcon}
Ebtesam Almazrouei, Hamza Alobeidli, Abdulaziz Alshamsi, Alessandro Cappelli,
  Ruxandra Cojocaru, Merouane Debbah, Etienne Goffinet, Daniel Heslow, Julien
  Launay, Quentin Malartic, Badreddine Noune, Baptiste Pannier, and Guilherme
  Penedo. 2023.
\newblock {Falcon-40B}: {A}n open large language model with state-of-the-art
  performance.
\newblock \url{https://huggingface.co/tiiuae/falcon-40b}.

\bibitem[{Anil et~al.(2023)Anil, Dai, Firat, Johnson, Lepikhin, Passos,
  Shakeri, Taropa, Bailey, Chen et~al.}]{anil2023palm}
Rohan Anil, Andrew~M Dai, Orhan Firat, Melvin Johnson, Dmitry Lepikhin,
  Alexandre Passos, Siamak Shakeri, Emanuel Taropa, Paige Bailey, Zhifeng Chen,
  et~al. 2023.
\newblock \href {http://arxiv.org/abs/arXiv:2305.10403} {Palm 2 technical
  report}.

\bibitem[{Austin et~al.(2021)Austin, Odena, Nye, Bosma, Michalewski, Dohan,
  Jiang, Cai, Terry, Le et~al.}]{austin2021program}
Jacob Austin, Augustus Odena, Maxwell Nye, Maarten Bosma, Henryk Michalewski,
  David Dohan, Ellen Jiang, Carrie Cai, Michael Terry, Quoc Le, et~al. 2021.
\newblock \href {http://arxiv.org/abs/arXiv:2108.07732} {Program synthesis with
  large language models}.

\bibitem[{Ba et~al.(2016)Ba, Kiros, and Hinton}]{ba2016layernorm}
Jimmy~Lei Ba, Jamie~Ryan Kiros, and Geoffrey~E. Hinton. 2016.
\newblock \href {http://arxiv.org/abs/1607.06450} {Layer normalization}.

\bibitem[{BAAI(2023)}]{baai2023Aquila2}
BAAI. 2023.
\newblock Aquila2 series proposed by {BAAI}.
\newblock \url{https://github.com/FlagAI-Open/Aquila2}.

\bibitem[{Bai et~al.(2023)Bai, Bai, Chu, Cui, Dang, Deng, Fan, Ge, Han, Huang
  et~al.}]{bai2023qwen}
Jinze Bai, Shuai Bai, Yunfei Chu, Zeyu Cui, Kai Dang, Xiaodong Deng, Yang Fan,
  Wenbin Ge, Yu~Han, Fei Huang, et~al. 2023.
\newblock \href {http://arxiv.org/abs/arXiv:2309.16609} {Qwen technical
  report}.

\bibitem[{Bai et~al.(2022)Bai, Kadavath, Kundu, Askell, Kernion, Jones, Chen,
  Goldie, Mirhoseini, McKinnon et~al.}]{bai2022constitutional}
Yuntao Bai, Saurav Kadavath, Sandipan Kundu, Amanda Askell, Jackson Kernion,
  Andy Jones, Anna Chen, Anna Goldie, Azalia Mirhoseini, Cameron McKinnon,
  et~al. 2022.
\newblock \href {http://arxiv.org/abs/arXiv:2212.08073} {Constitutional {AI}:
  {H}armlessness from {AI} feedback}.

\bibitem[{Baichuan(2023)}]{baichuan1}
Baichuan. 2023.
\newblock A large-scale 7{B} pretraining language model developed by baichuan
  {I}nc.
\newblock \url{https://github.com/baichuan-inc/Baichuan-7B}.

\bibitem[{Brown et~al.(2020)Brown, Mann, Ryder, Subbiah, Kaplan, Dhariwal,
  Neelakantan, Shyam, Sastry, Askell, Agarwal, Herbert-Voss, Krueger, Henighan,
  Child, Ramesh, Ziegler, Wu, Winter, Hesse, Chen, Sigler, Litwin, Gray, Chess,
  Clark, Berner, McCandlish, Radford, Sutskever, and
  Amodei}]{brown2020language}
Tom Brown, Benjamin Mann, Nick Ryder, Melanie Subbiah, Jared~D Kaplan, Prafulla
  Dhariwal, Arvind Neelakantan, Pranav Shyam, Girish Sastry, Amanda Askell,
  Sandhini Agarwal, Ariel Herbert-Voss, Gretchen Krueger, Tom Henighan, Rewon
  Child, Aditya Ramesh, Daniel Ziegler, Jeffrey Wu, Clemens Winter, Chris
  Hesse, Mark Chen, Eric Sigler, Mateusz Litwin, Scott Gray, Benjamin Chess,
  Jack Clark, Christopher Berner, Sam McCandlish, Alec Radford, Ilya Sutskever,
  and Dario Amodei. 2020.
\newblock Language models are few-shot learners.
\newblock In \emph{Advances in Neural Information Processing Systems}.

\bibitem[{Chen et~al.(2021)Chen, Tworek, Jun, Yuan, Pinto, Kaplan, Edwards,
  Burda, Joseph, Brockman et~al.}]{chen2021evaluating}
Mark Chen, Jerry Tworek, Heewoo Jun, Qiming Yuan, Henrique Ponde de~Oliveira
  Pinto, Jared Kaplan, Harri Edwards, Yuri Burda, Nicholas Joseph, Greg
  Brockman, et~al. 2021.
\newblock \href {http://arxiv.org/abs/arXiv:2107.03374} {Evaluating large
  language models trained on code}.

\bibitem[{Cobbe et~al.(2021)Cobbe, Kosaraju, Bavarian, Chen, Jun, Kaiser,
  Plappert, Tworek, Hilton, Nakano et~al.}]{cobbe2021training}
Karl Cobbe, Vineet Kosaraju, Mohammad Bavarian, Mark Chen, Heewoo Jun, Lukasz
  Kaiser, Matthias Plappert, Jerry Tworek, Jacob Hilton, Reiichiro Nakano,
  et~al. 2021.
\newblock \href {http://arxiv.org/abs/arXiv:2110.14168} {Training verifiers to
  solve math word problems}.

\bibitem[{Computer(2023)}]{redpajamav2}
Together Computer. 2023.
\newblock Red{P}ajama: {A}n open dataset for training large language models.
\newblock \url{https://github.com/togethercomputer/RedPajama-Data}.

\bibitem[{Cui et~al.(2023)Cui, Yang, and Yao}]{cui2023chinesealpaca2}
Yiming Cui, Ziqing Yang, and Xin Yao. 2023.
\newblock \href {http://arxiv.org/abs/arXiv:2304.08177} {Efficient and
  effective text encoding for {C}hinese {LLaMA} and {A}lpaca}.

\bibitem[{Dao(2023)}]{dao2023flashattention2}
Tri Dao. 2023.
\newblock \href {http://arxiv.org/abs/arXiv:2307.08691} {Flashattention-2:
  {F}aster attention with better parallelism and work partitioning}.

\bibitem[{Hendrycks et~al.(2021{\natexlab{a}})Hendrycks, Burns, Basart, Zou,
  Mazeika, Song, and Steinhardt}]{hendrycks2020measuring}
Dan Hendrycks, Collin Burns, Steven Basart, Andy Zou, Mantas Mazeika, Dawn
  Song, and Jacob Steinhardt. 2021{\natexlab{a}}.
\newblock Measuring massive multitask language understanding.
\newblock In \emph{International Conference on Learning Representations}.

\bibitem[{Hendrycks et~al.(2021{\natexlab{b}})Hendrycks, Burns, Kadavath,
  Arora, Basart, Tang, Song, and Steinhardt}]{hendrycks2021measuring}
Dan Hendrycks, Collin Burns, Saurav Kadavath, Akul Arora, Steven Basart, Eric
  Tang, Dawn Song, and Jacob Steinhardt. 2021{\natexlab{b}}.
\newblock Measuring mathematical problem solving with the {MATH} dataset.
\newblock In \emph{Conference on Neural Information Processing Systems Track on
  Datasets and Benchmarks}.

\bibitem[{Huang et~al.(2023)Huang, Bai, Zhu, Zhang, Zhang, Su, Liu, Lv, Zhang,
  Lei et~al.}]{huang2023c}
Yuzhen Huang, Yuzhuo Bai, Zhihao Zhu, Junlei Zhang, Jinghan Zhang, Tangjun Su,
  Junteng Liu, Chuancheng Lv, Yikai Zhang, Jiayi Lei, et~al. 2023.
\newblock \href {http://arxiv.org/abs/arXiv:2305.08322} {C-{E}val: {A}
  multi-level multi-discipline chinese evaluation suite for foundation models}.

\bibitem[{InternLM(2023)}]{internlm2023internlm}
InternLM. 2023.
\newblock Intern{LM}: {A} multilingual language model with progressively
  enhanced capabilities.
\newblock \url{https://github.com/InternLM/InternLM}.

\bibitem[{Kaddour et~al.(2023)Kaddour, Harris, Mozes, Bradley, Raileanu, and
  McHardy}]{kaddour2023challenges}
Jean Kaddour, Joshua Harris, Maximilian Mozes, Herbie Bradley, Roberta
  Raileanu, and Robert McHardy. 2023.
\newblock \href {http://arxiv.org/abs/arXiv:2307.10169} {Challenges and
  applications of large language models}.

\bibitem[{Kingma and Ba(2015)}]{diederik2015adam}
Diederik~P. Kingma and Jimmy Ba. 2015.
\newblock \href {http://arxiv.org/abs/1412.6980} {Adam: {A} method for
  stochastic optimization}.
\newblock In \emph{3rd International Conference on Learning Representations,
  {ICLR} 2015, San Diego, CA, USA, May 7-9, 2015, Conference Track
  Proceedings}.

\bibitem[{Kudo and Richardson(2018)}]{kudo2018sentencepiece}
Taku Kudo and John Richardson. 2018.
\newblock Sentence{P}iece: {A} simple and language independent subword
  tokenizer and detokenizer for neural text processing.
\newblock In \emph{Conference on Empirical Methods in Natural Language
  Processing: System Demonstrations}, pages 66--71.

\bibitem[{Lefaudeux et~al.(2022)Lefaudeux, Massa, Liskovich, Xiong, Caggiano,
  Naren, Xu, Hu, Tintore, Zhang, Labatut, and Haziza}]{xFormers2022}
Benjamin Lefaudeux, Francisco Massa, Diana Liskovich, Wenhan Xiong, Vittorio
  Caggiano, Sean Naren, Min Xu, Jieru Hu, Marta Tintore, Susan Zhang, Patrick
  Labatut, and Daniel Haziza. 2022.
\newblock xformers: A modular and hackable transformer modelling library.
\newblock \url{https://github.com/facebookresearch/xformers}.

\bibitem[{Li et~al.(2023)Li, Zhang, Koto, Yang, Zhao, Gong, Duan, and
  Baldwin}]{li2023cmmlu}
Haonan Li, Yixuan Zhang, Fajri Koto, Yifei Yang, Hai Zhao, Yeyun Gong, Nan
  Duan, and Timothy Baldwin. 2023.
\newblock \href {http://arxiv.org/abs/arXiv:2306.09212} {{CMMLU}: {M}easuring
  massive multitask language understanding in chinese}.

\bibitem[{Lightman et~al.(2023)Lightman, Kosaraju, Burda, Edwards, Baker, Lee,
  Leike, Schulman, Sutskever, and Cobbe}]{lightman2023let}
Hunter Lightman, Vineet Kosaraju, Yura Burda, Harri Edwards, Bowen Baker, Teddy
  Lee, Jan Leike, John Schulman, Ilya Sutskever, and Karl Cobbe. 2023.
\newblock \href {http://arxiv.org/abs/arXiv:2305.20050} {Let's verify step by
  step}.

\bibitem[{Loshchilov and Hutter(2017)}]{Loshchilov2017DecoupledWD}
Ilya Loshchilov and Frank Hutter. 2017.
\newblock \href {https://api.semanticscholar.org/CorpusID:53592270} {Decoupled
  weight decay regularization}.
\newblock In \emph{International Conference on Learning Representations}.

\bibitem[{MosaicML et~al.(2023)}]{mn2023introducing}
MosaicML et~al. 2023.
\newblock {MPT-30B}: {R}aising the bar for open-source foundation models.
\newblock \url{https://www.mosaicml.com/blog/mpt-30b}.

\bibitem[{OpenCompass(2023)}]{2023opencompass}
OpenCompass. 2023.
\newblock Open{C}ompass: {A} universal evaluation platform for foundation
  models.
\newblock \url{https://github.com/open-compass/opencompass}.

\bibitem[{{Ortiz Su{\'a}rez} et~al.(2019){Ortiz Su{\'a}rez}, Sagot, and
  Romary}]{OSCAR}
Pedro~Javier {Ortiz Su{\'a}rez}, Beno{\^i}t Sagot, and Laurent Romary. 2019.
\newblock Asynchronous pipelines for processing huge corpora on medium to low
  resource infrastructures.
\newblock In \emph{Workshop on the Challenges in the Management of Large
  Corpora}.

\bibitem[{Ouyang et~al.(2022)Ouyang, Wu, Jiang, Almeida, Wainwright, Mishkin,
  Zhang, Agarwal, Slama, Ray et~al.}]{ouyang2022training}
Long Ouyang, Jeffrey Wu, Xu~Jiang, Diogo Almeida, Carroll Wainwright, Pamela
  Mishkin, Chong Zhang, Sandhini Agarwal, Katarina Slama, Alex Ray, et~al.
  2022.
\newblock Training language models to follow instructions with human feedback.
\newblock \emph{Advances in Neural Information Processing Systems}.

\bibitem[{Penedo et~al.(2023)Penedo, Malartic, Hesslow, Cojocaru, Cappelli,
  Alobeidli, Pannier, Almazrouei, and Launay}]{penedo2023refinedweb}
Guilherme Penedo, Quentin Malartic, Daniel Hesslow, Ruxandra Cojocaru,
  Alessandro Cappelli, Hamza Alobeidli, Baptiste Pannier, Ebtesam Almazrouei,
  and Julien Launay. 2023.
\newblock \href {http://arxiv.org/abs/arXiv:2306.01116} {The {R}efined{W}eb
  dataset for {F}alcon {LLM}: {O}utperforming curated corpora with web data,
  and web data only}.

\bibitem[{Peng et~al.(2023)Peng, Quesnelle, Fan, and Shippole}]{peng2023yarn}
Bowen Peng, Jeffrey Quesnelle, Honglu Fan, and Enrico Shippole. 2023.
\newblock \href {http://arxiv.org/abs/arXiv:2309.00071} {Yarn: {E}fficient
  context window extension of large language models}.

\bibitem[{Press et~al.(2022)Press, Smith, and Lewis}]{alibi}
Ofir Press, Noah Smith, and Mike Lewis. 2022.
\newblock Train short, test long: Attention with linear biases enables input
  length extrapolation.
\newblock In \emph{International Conference on Learning Representations}.

\bibitem[{Rajbhandari et~al.(2020)Rajbhandari, Rasley, Ruwase, and
  He}]{rajbhandari2020zero}
Samyam Rajbhandari, Jeff Rasley, Olatunji Ruwase, and Yuxiong He. 2020.
\newblock Zero: Memory optimizations toward training trillion parameter models.
\newblock In \emph{SC20: International Conference for High Performance
  Computing, Networking, Storage and Analysis}, pages 1--16. IEEE.

\bibitem[{Schulman et~al.(2017)Schulman, Wolski, Dhariwal, Radford, and
  Klimov}]{schulman2017PPO}
John Schulman, Filip Wolski, Prafulla Dhariwal, Alec Radford, and Oleg Klimov.
  2017.
\newblock \href {http://arxiv.org/abs/1707.06347} {Proximal policy optimization
  algorithms}.

\bibitem[{Shazeer(2019)}]{Shazeer2019FastTD}
Noam Shazeer. 2019.
\newblock \href {http://arxiv.org/abs/arXiv:1911.02150} {Fast transformer
  decoding: {O}ne write-head is all you need}.

\bibitem[{Shazeer(2020)}]{shazeer2020glu}
Noam Shazeer. 2020.
\newblock \href {http://arxiv.org/abs/arXiv:2002.05202} {Glu variants improve
  transformer}.

\bibitem[{Shen et~al.(2023)Shen, Tao, Ma, Neiswanger, Liu, Wang, Tan, Hestness,
  Vassilieva, Soboleva, and Xing}]{shen2023slimpajama}
Zhiqiang Shen, Tianhua Tao, Liqun Ma, Willie Neiswanger, Zhengzhong Liu, Hongyi
  Wang, Bowen Tan, Joel Hestness, Natalia Vassilieva, Daria Soboleva, and Eric
  Xing. 2023.
\newblock \href {http://arxiv.org/abs/2309.10818} {{SlimPajama-DC}:
  {U}nderstanding data combinations for {LLM} training}.

\bibitem[{Shibatay et~al.(1999)Shibatay, Kiday, Fukamachiz, Takeday,
  Shinoharay, Shinoharaz, and Arikaway}]{shibata1999BPE:BytePairencoding}
Yusuke Shibatay, Takuya Kiday, Shuichi Fukamachiz, Masayuki Takeday, Ayumi
  Shinoharay, Takeshi Shinoharaz, and Setsuo Arikaway. 1999.
\newblock Byte pair encoding: {A} text compression scheme that accelerates
  pattern matching.
\newblock Technical Report DOI-TR-161, Department of Informatics, Kyushu
  University.

\bibitem[{Su et~al.(2023)Su, Ahmed, Lu, Pan, Bo, and Liu}]{su2023roformer}
Jianlin Su, Murtadha Ahmed, Yu~Lu, Shengfeng Pan, Wen Bo, and Yunfeng Liu.
  2023.
\newblock Roformer: {E}nhanced transformer with rotary position embedding.
\newblock \emph{Neurocomputing}, page 127063.

\bibitem[{Suzgun et~al.(2022)Suzgun, Scales, Sch{\"a}rli, Gehrmann, Tay, Chung,
  Chowdhery, Le, Chi, Zhou et~al.}]{suzgun2022challenging}
Mirac Suzgun, Nathan Scales, Nathanael Sch{\"a}rli, Sebastian Gehrmann, Yi~Tay,
  Hyung~Won Chung, Aakanksha Chowdhery, Quoc~V Le, Ed~H Chi, Denny Zhou, et~al.
  2022.
\newblock \href {http://arxiv.org/abs/arXiv:2210.09261} {Challenging big-bench
  tasks and whether chain-of-thought can solve them}.

\bibitem[{Touvron et~al.(2023{\natexlab{a}})Touvron, Lavril, Izacard, Martinet,
  Lachaux, Lacroix, Rozi{\`e}re, Goyal, Hambro, Azhar
  et~al.}]{touvron2023llama}
Hugo Touvron, Thibaut Lavril, Gautier Izacard, Xavier Martinet, Marie-Anne
  Lachaux, Timoth{\'e}e Lacroix, Baptiste Rozi{\`e}re, Naman Goyal, Eric
  Hambro, Faisal Azhar, et~al. 2023{\natexlab{a}}.
\newblock \href {http://arxiv.org/abs/2302.13971} {{LLaMA}: {O}pen and
  efficient foundation language models}.

\bibitem[{Touvron et~al.(2023{\natexlab{b}})Touvron, Martin, Stone, Albert,
  Almahairi, Babaei, Bashlykov, Batra, Bhargava, Bhosale
  et~al.}]{touvron2023llama2}
Hugo Touvron, Louis Martin, Kevin Stone, Peter Albert, Amjad Almahairi, Yasmine
  Babaei, Nikolay Bashlykov, Soumya Batra, Prajjwal Bhargava, Shruti Bhosale,
  et~al. 2023{\natexlab{b}}.
\newblock \href {http://arxiv.org/abs/2307.09288} {{LLaMA} 2: {O}pen foundation
  and fine-tuned chat models}.

\bibitem[{Vaswani et~al.(2017)Vaswani, Shazeer, Parmar, Uszkoreit, Jones,
  Gomez, Kaiser, and Polosukhin}]{vaswani2017attention}
Ashish Vaswani, Noam Shazeer, Niki Parmar, Jakob Uszkoreit, Llion Jones,
  Aidan~N Gomez, {\L}ukasz Kaiser, and Illia Polosukhin. 2017.
\newblock Attention is all you need.
\newblock In \emph{Advances in Neural Information Processing Systems}.

\bibitem[{Workshop et~al.(2022)Workshop, Scao, Fan, Akiki, Pavlick, Ili{\'c},
  Hesslow, Castagn{\'e}, Luccioni, Yvon et~al.}]{workshop2022bloom}
BigScience Workshop, Teven~Le Scao, Angela Fan, Christopher Akiki, Ellie
  Pavlick, Suzana Ili{\'c}, Daniel Hesslow, Roman Castagn{\'e}, Alexandra~Sasha
  Luccioni, Fran{\c{c}}ois Yvon, et~al. 2022.
\newblock \href {http://arxiv.org/abs/arXiv:2211.05100} {Bloom: {A}
  176{B}-parameter open-access multilingual language model}.

\bibitem[{XVERSE(2023)}]{xverse2023xverse}
XVERSE. 2023.
\newblock {XVERSE-13B}: {A} multilingual large language model developed by
  {XVERSE} {T}echnology {I}nc.
\newblock \url{https://github.com/xverse-ai/XVERSE-13B}.

\bibitem[{Yang et~al.(2023)Yang, Xiao, Wang, Zhang, Yin, Lv, Pan, Wang, Yan,
  Yang et~al.}]{baichuan2}
Aiyuan Yang, Bin Xiao, Bingning Wang, Borong Zhang, Chao Yin, Chenxu Lv,
  Da~Pan, Dian Wang, Dong Yan, Fan Yang, et~al. 2023.
\newblock \href {http://arxiv.org/abs/2309.10305} {Baichuan 2: {O}pen
  large-scale language models}.

\bibitem[{Zeng et~al.(2023)Zeng, Liu, Du, Wang, Lai, Ding, Yang, Xu, Zheng, Xia
  et~al.}]{zeng2022chatglm}
Aohan Zeng, Xiao Liu, Zhengxiao Du, Zihan Wang, Hanyu Lai, Ming Ding, Zhuoyi
  Yang, Yifan Xu, Wendi Zheng, Xiao Xia, et~al. 2023.
\newblock {GLM-130B}: {A}n open bilingual pre-trained model.
\newblock In \emph{International Conference on Learning Representations}.

\bibitem[{Zhang and Sennrich(2019)}]{zhang2019RMS}
Biao Zhang and Rico Sennrich. 2019.
\newblock Root mean square layer normalization.
\newblock In \emph{Advances in Neural Information Processing Systems}.

\bibitem[{Zhang et~al.(2023)Zhang, Li, Zong, Ying, He, and
  Qiu}]{zhang2023evaluating}
Xiaotian Zhang, Chunyang Li, Yi~Zong, Zhengyu Ying, Liang He, and Xipeng Qiu.
  2023.
\newblock \href {http://arxiv.org/abs/arXiv:2305.12474} {Evaluating the
  performance of large language models on {GAOKAO} benchmark}.

\bibitem[{Zhong et~al.(2023)Zhong, Cui, Guo, Liang, Lu, Wang, Saied, Chen, and
  Duan}]{zhong2023agieval}
Wanjun Zhong, Ruixiang Cui, Yiduo Guo, Yaobo Liang, Shuai Lu, Yanlin Wang, Amin
  Saied, Weizhu Chen, and Nan Duan. 2023.
\newblock \href {http://arxiv.org/abs/arXiv:2304.06364} {{AGIE}val: {A}
  human-centric benchmark for evaluating foundation models}.

\end{thebibliography}

\end{document}